\documentclass{article}

\usepackage[english]{babel}
\usepackage[utf8]{inputenc}
\usepackage{amsmath}
\usepackage{amsfonts}
\usepackage{amssymb}
\usepackage{graphicx}
\graphicspath{{figures/}}
\usepackage{color}
\usepackage{hyperref}
\usepackage{xspace}
\usepackage{xifthen}
\usepackage{multicol}
\usepackage{mathtools}
\usepackage{dsfont}
\usepackage{tikz}
\usepackage{cleveref}

\usepackage{xspace}
\usepackage{xifthen}
\usepackage{mathtools}
\usepackage{dsfont}
\usepackage{my_notations}

\newcommand{\pkg}[1]{\texttt{#1}}

\newtheorem{definition}{Definition}

\newcommand{\Shsici}[1][]{\ifthenelse{\isempty{#1}}{\enm{\text{S}^{\text{HSIC}}_{q,\textbf{T}}}}{\enm{\text{S}^{\text{HSIC}}_{#1}}}}
\newcommand*{\Aset}{\enm{\mathcal A}}
\newcommand*{\Acomp}{\enm{\stcomp{\Aset}}}

\usepackage{lipsum}
\usepackage{epstopdf}
\usepackage{algorithmic}
\ifpdf
  \DeclareGraphicsExtensions{.eps,.pdf,.png,.jpg}
\else
  \DeclareGraphicsExtensions{.eps}
\fi

\newcommand{\TheTitle}{Global sensitivity analysis for optimization with variable selection}
\newcommand{\TheAuthors}{A. Spagnol, R. Le Riche, and S. Da Veiga}

\usepackage{amsopn}

\ifpdf
\hypersetup{
  pdftitle={\TheTitle},
  pdfauthor={\TheAuthors}
}
\fi

\let\svthefootnote\thefootnote
\newcommand\blankfootnote[1]{%
  \let\thefootnote\relax\footnotetext{#1}%
  \let\thefootnote\svthefootnote%
}
\let\svfootnote\footnote
\renewcommand\footnote[2][?]{%
  \if\relax#1\relax%
    \blankfootnote{#2}%
  \else%
    \if?#1\svfootnote{#2}\else\svfootnote[#1]{#2}\fi%
  \fi
}

\usepackage[normalem]{ulem}
\usepackage{collab_tex}
\usepackage{tikz}

\begin{document}

\title{Global sensitivity analysis for optimization with variable selection}
\author{Adrien Spagnol$^{1,3}$\quad Rodolphe Le Riche$^{2,1}$ \quad Sebastien Da Veiga$^{3}$}
\footnote[]{\small $^1$Ecole Nationale Superieure des Mines de Saint-Etienne, F-42023 Saint-Etienne, France,~\small {\tt leriche@emse.fr}}
\footnote[]{\small $^2$CNRS LIMOS UMR 6158, France}
\footnote[]{\small $^3$Safran Tech, Safran SA, Magny-Les-Hameaux, France,  \small{\tt adrien.spagnol@safrangroup.fr}, ~\small{\tt sebastien.da-veiga@safrangroup.com}
}
\date{} 
\maketitle
\bf This work is currently being submitted and under reviews for  the SIAM/ASA Journal on Uncertainty Quantification \rm
%
\begin{abstract}
The optimization of high dimensional functions is a key issue in engineering problems but it frequently comes at a cost that is not acceptable since it usually involves a complex and expensive computer code. 
Engineers often overcome this limitation by first identifying which parameters drive the most the function variations: non-influential variables are set to a fixed value and the optimization procedure is carried out with the remaining influential variables. Such variable selection is performed through influence measures that are meaningful for regression problems. However it does not account for the specific structure of optimization problems where we would like to identify which variables most lead to constraints satisfaction and low values of the objective function. \\
In this paper, we propose a new sensitivity analysis that accounts for the specific aspects of optimization problems. In particular, we introduce an influence measure based on the Hilbert-Schmidt Independence Criterion to characterize whether a design variable matters to reach low values of the objective function and to satisfy the constraints. This sensitivity measure makes it possible to sort the inputs and reduce the problem dimension. We compare a random and a greedy strategies to set the values of the non-influential variables before conducting a local optimization. Applications to several test-cases show that this variable selection and the greedy strategy significantly reduce the number of function evaluations at a limited cost in terms of solution performance.
\end{abstract}



\section{Introduction}
A common engineering practice is to look for performance by optimizing models of the considered system. However, these models often have a significant numerical cost that can, in extreme cases, reach days for a single simulation. Another difficulty for optimization appears when models rely on a large numbers of inputs whose direct importance on the output is difficult to apprehend. In order to identify the parameters of the model that most affect the performance and reduce the number of parameters involved in the optimization, one can resort to global sensitivity analysis.

Sensitivity analysis has emerged as a set of methodologies to capture the impact of each input on the model output variability \cite{iooss2015review}. By doing so, one can assess the relevance of each input, reduce the number of dimensions, and have a better understanding of a model. Sensitivity analysis methods can be sorted out into two main groups: regression-based and variance-based methods.

Because regression-based methods fail to properly represent the model sensitivity when it is not monotonic \cite{iooss2015review}, variance techniques have gained in popularity since Sobol's work \cite{sobol1993sensitivity}. These methods rely on the decomposition of the variance of the output as a sum of contributions of each input variable or combinations thereof, hence their name, ANOVA techniques (for ``Analysis Of Variance''). The indices established by Sobol estimate each input variable weight in the variance of the model response and are usually evaluated by raw Monte Carlo simulations. 

Even though the variance-based  global sensitivity estimators are commonly used in optimization, they suffer from certain drawbacks. First, because of their formulation, they only capture the effect of a variable on the conditional mean of the output, which may fail to give a proper estimation of the importance of each input when studying quantities such as quantiles or probabilities of exceedance. This has lead to the recent development of new sensitivity indices \cite{fort2016new} that can handle cases unrelated to a variance criterion. Second, despite the recent work of \cite{gamboa2014sensitivity}, these indices do not easily extend to multivariate outputs. 

Recently a new class of sensitivity indices based on dependence measures has received some attention: these approaches, such as the popular $\delta$-sensitivity measure of \cite{borgonovo2007new}, measure how much the output distribution differs from the conditional output distribution given a specific input or group of inputs. Sensitivity analysis relying on the Hilbert-Schmidt Independence Criterion (HSIC), introduced by \cite{gretton2005measuring}, is a kernel-based approach showing promising results for sensitivity analysis since it measures the impact of a variable on the full distribution of the system output. 

The goals of this paper are, first, to propose a generalization of sensitivity analysis for optimization problems and, second, to study how sensitivity analysis can contribute to reducing optimization cost. Capturing which inputs are important in order to reach an optimal model performance is a key information. Additionally, some variables that appear to have a negligible impact on the overall performance may be useful specifically near the optimum output value. In this paper, we propose three different modifications of the model output that allow characterizing three relationships between the high-performance regions and the entire the design space.
Then, an optimization strategy is defined and studied where, based on the introduced sensitivity measures, variables are freezed at specific values and the search is carried out in a lower dimensional space. 

\Cref{sec:vbasedSA} introduces variance-based sensitivity analysis for both scalar and vector outputs, and proposes and illustrates modifications for optimization problems. \Cref{sec:kbasedSA} concentrates on kernel-based sensitivity analysis and highlights interesting properties of the aforementioned modifications, leading to a new HSIC-IT sensitivity measure. Finally, \Cref{sec:examples} describes two ways to use the HSIC-IT measure in optimization, reports two constrained test examples and discusses the results in terms of computational efficiency and solution accuracy.

\section{Variance-based sensitivity analysis}
\label{sec:vbasedSA}

\subsection{Functional variance decomposition and Sobol Indices}
Since their introduction \cite{sobol1993sensitivity}, the Sobol indices have become the most widely used variance-based sensitivity measures in industrial applications. 
Considering a model $Y = f(X)$ and assuming that $f(X) \in L^2(\mathbb{R})$, the functional ANOVA decomposition of $f(X)$ (see \cite{efron1981jackknife}, \cite{hoeffding1948class}, \cite{sobol1993sensitivity}) is the sum of functions of increasing dimension

\begin{equation} \label{eq:ANOVA}
\begin{aligned}
f(X) & = f_0 + \sum^d_{i=1}{f_i(X_i)} + \sum_{1 \leq i \le j \leq d}{f_{i,j}(X_i, X_j)} + \cdots + f_{1,2, \ldots,d}(\textbf{X}) \\
  & = \sum_{u \subset \{1,\ldots,d\}}{f_u(X_u)},
\end{aligned}
\end{equation}
where $f_0 = \mathbb{E}[f(X)]$, $f_i(X_i) = \mathbb{E}[f(X)|X_i] - f_0$ and $f_u(X_u) = \mathbb{E}[f(X)|X_u] - \sum_{v \subset u}{f_v(X_v)}$ with $X_u = (X_i)_{i \in u}$, for any $u \subset \{1,\ldots,d\}$, i.e. all the possible subset combinations without repetitions. This decomposition exists and is unique under conditions. From \eqref{eq:ANOVA}, we can derive the Sobol indices, or variance-based indices, for any $u \subset \{1,\ldots,d\}$, as

\begin{equation} \label{eq:Sobol_ind}
S_u = \frac{\text{Var}[f_u(X_u)]}{\text{Var}[f(X)]}.
\end{equation}

Those indices express the share in variance of $Y$ of an input or a group of inputs and they sum to one. The total number of indices rises to $2^d - 1$ thus, indices of order higher than two are usually not evaluated to save computational time and because they become difficult to interprete. The \emph{total} Sobol indices have been introduced in \cite{homma1996importance} as
\begin{equation}
S^T_u = \sum_{\substack{v \subset \{1,\ldots,d\} \\ {v \supset u}}} S_v.
\end{equation}
The total Sobol index $S^T_u$ measures the effect of $X_u$ and all of its interactions with the other $X_j$. Most of the time, when $d$ becomes large, only the first and total indices are computed. 

Several techniques, usually based on Monte-Carlo estimations, have been devised to compute those sensitivity indices efficiently, see \cite{saltelli2010variance}.

\subsection{Applying variance-based sensitivity analysis to optimization problems} \label{ssec:formulation}

The main goal of our study is to minimize an objective function $f(X)$ under $m$ inequality constraints $g_l(X)$,
\begin{equation}
\begin{aligned}
& \underset{X \in \mathbb{R}^d}{\text{minimize}} & & f(X),\\
& \text{subject to} & & g_l(X) \leq 0, \; l = 1, \ldots, m.
\end{aligned}
\label{eq-opt_original}
\end{equation}
Traditionally, the Sobol indices of the inputs are computed and the inputs whose indices are close to zero are set to a fixed value, often the nominal value (when defined). 
Having fewer variables in the optimization problem implies a smaller search volume, hence fewer calls to the model to reach an optimum. However, since the low-impacting inputs are fixed, some of the ability to tune the output is lost in the simplification and the optimum of the modified problem might differ from the real global optimum. 
In the case of constrained optimization, the Sobol indices are also computed on these new outputs or on an aggregation of the constraints and the objective functions using the multiple output version of Sobol indices \cite{gamboa2014sensitivity}. 

\begin{figure} [h!]
\centering
   \begin{minipage}{.5\linewidth}
   	  \centering
      \includegraphics[width=0.8\linewidth]{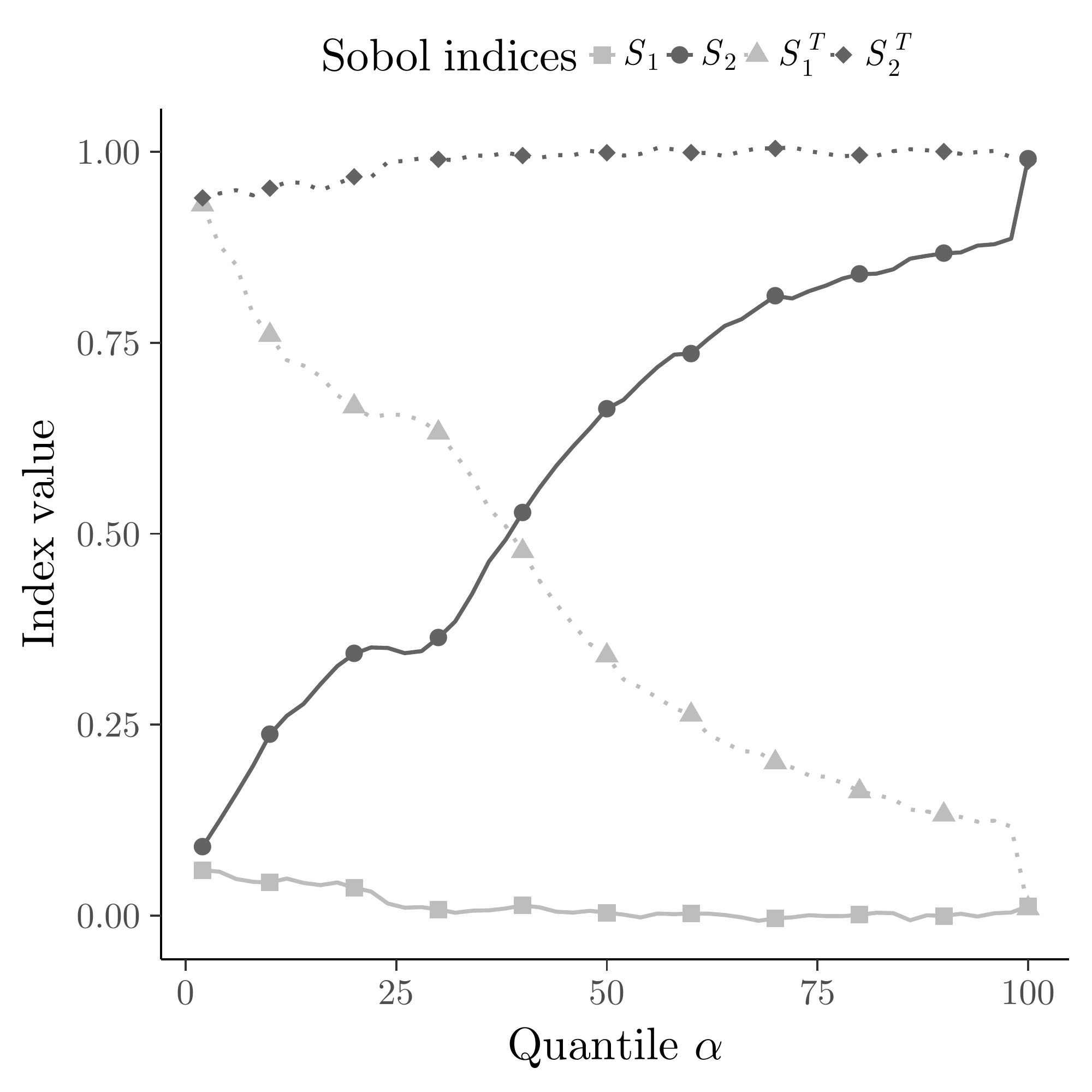}
   \end{minipage}%
   \begin{minipage}{.5\linewidth}
   	  \centering
      \includegraphics[width=0.8\textwidth]{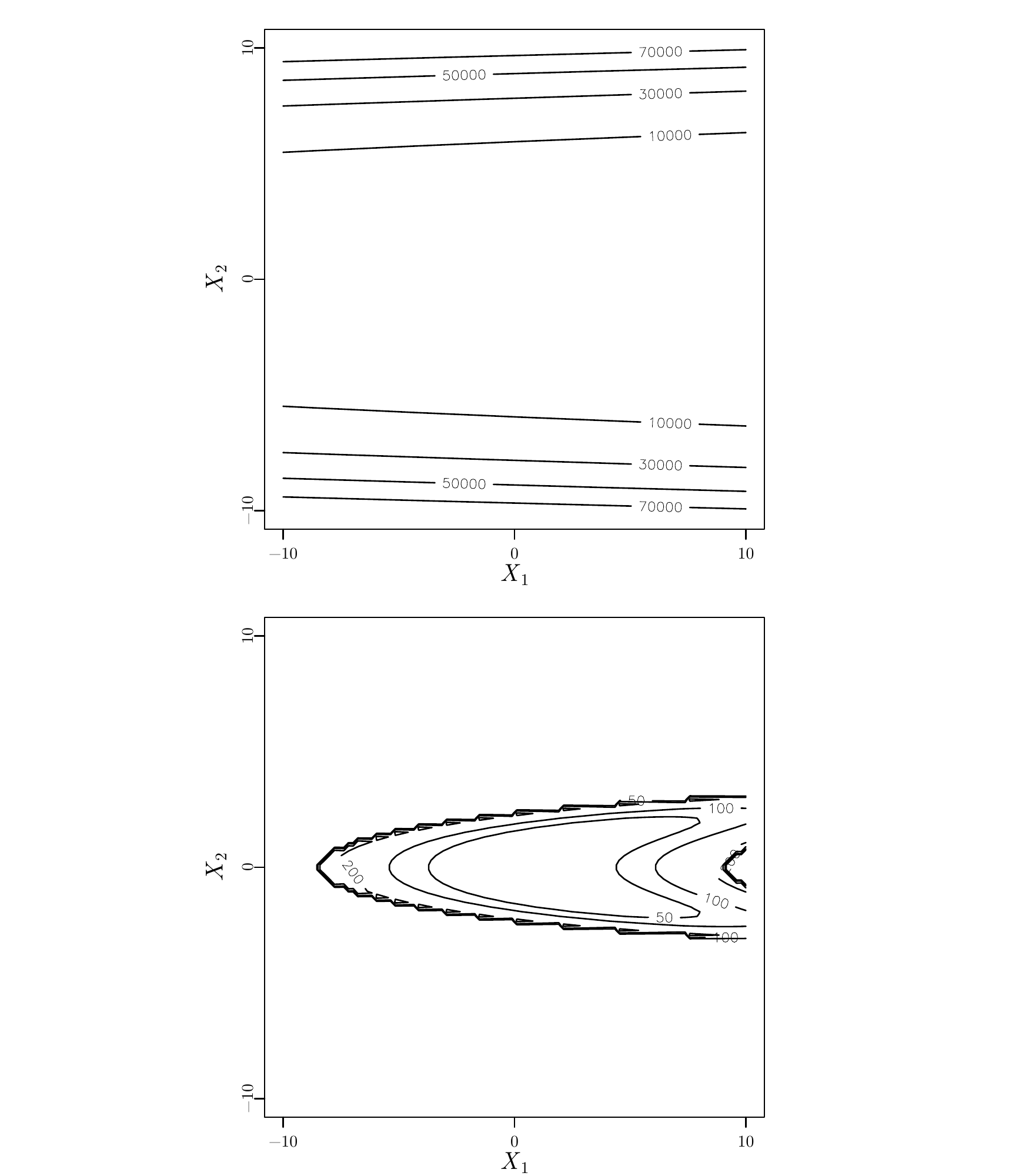}
   \end{minipage}
   
\caption{\textbf{Left:} Evolution of $S_1$ and $S_1^T$, resp. $S_2$ and $S_2^T$, with respect to the $\alpha$-quantile for the Dixon-Price function \eqref{eq:DixonPr} using the zero-thresholding. \textbf{Upper-right:} contour of the function, \textbf{bottom-right:} contour of the thresholded function  $\mathcal{D} = \{f(X) \leq q_{20\%}\}$. The contour lines on the right-hand side no longer correspond to an ellipse aligned with the reference axes, there is a change of curvature associated to a Sobol dependency between the variable.}
\label{fig:Sobol_Indic_DixonPr}
\end{figure}

Yet, this approach does not correspond to what we are aiming for: Sobol indices give information on the influence of an input in the full design domain, while we are interested in finding which variables are important, \textit{i)} in order to  respect the constraints and have a sufficiently low objective function, and \textit{ii)} when the constraints are satisfied and the objective function is sufficiently low. We use the two-dimensional Dixon-Price function as an example: 
\begin{equation} \label{eq:DixonPr}
f(X) = (X_1 - 1)^2 + 2(2X_2^2 - X_1)^2,
\end{equation}
with $X_i \sim \mathcal{U}[-10,10]$, for $i = \{1,2\}$. A sensitivity analysis of the function gives a first-order index equal to 0 for $X_1$ while it is close to 1 for $X_2$. In the light of this analysis, the first input appears as negligible and can be set to a fixed value, such as its mean value $\mu_{X_1} = 0$. But, doing so makes it impossible to find the global optimum $X^* = [1,1]$ because of the loss in fine tuning. 
As it can be seen in the right plots of \cref{fig:Sobol_Indic_DixonPr}, low values of the Dixon-Price function have skewed contour lines showing that what matters to find the global minimum  is the interaction of both variables. 
To assess the effect of the inputs on a function to optimize, \cite{fort2016new} have defined the \textit{goal-oriented sensitivity indices}, which quantify the importance of each input $X_i$ based on a quantity of interest to estimate. They show that Sobol indices are a specific case of their new indices, since they characterize the sensitivity of an input with respect to the mean. 

In the current study, we derive sensitivity indices adapted to optimization by three thresholding transformations of the output $f(X)$ and by performing sensitivity analysis on the modified output which is written $Z$. 
We define $\mathcal{D}$, the sublevel set where the objective is below a given threshold $q$ and the constraints are respected up to $\textbf{T}$, $\mathcal{D} = \{X \in \mathbb{R}^d, \textbf{g}(X) \leq \textbf{T} \text{ and } f(X) \leq q\}$, with $\textbf{T} \in \mathbb{R}^{m,+}$ and $q \in \mathbb{R}$. 
The threshold $\textbf{T}$ relax the constraints when finding a feasible point is too difficult. In this paper, the $\textbf{T}$ values are all similar and chosen in order to have a sufficient number of feasible points for the sensitivity indices computation, i.e. a few hundreds. The threshold $q$ that contributes to the definition of $\mathcal{D}$ is a quantile $q_\alpha$ of the objective function $f(\cdot)$. Low quantiles ensure that we are looking at values of the output close to the best observations.

We consider the following output transformations based on thresholding:
\begin{enumerate}
\item Zero-thresholding: $Z = f(X) \times \mathds{1}_{X \in \mathcal{D}}$; 

\item Conditional-thresholding: $Z = f(X) | (X \in \mathcal{D})$; ;

\item Indicator-thresholding: $Z = \mathds{1}_{X \in \mathcal{D}}$. 

\end{enumerate}
where $\mathds{1}$ is the indicator function. 

We now discuss these different thresholdings.

\paragraph{Zero-thresholding\protect}

Recalling the previous example of the two-dimen-sional Dixon-Price function, we define\footnotemark[1]\footnotetext[1]{As a special case for the more general $C$ constant-thresholding, $Z = f(X) \times \mathds{1}_{X \in \mathcal{D}} + C \times \mathds{1}_{X \notin \mathcal{D}}$} $Z = f(X) \times \mathds{1}_{f(X) \leq q_\alpha}$ and compute the first and total order Sobol indices of $Z$ with respect to the value of $\alpha$, see the left side of \cref{fig:Sobol_Indic_DixonPr}. In 2 dimensions, $S_1^T = S_1 + S_{12}$, resp. $S_2^T = S_2 + S_{12}$, where $S_{12}$ is the second-order index which characterizes the effect of $X_1$ and $X_2$ varying simultaneously. In that case, when $\alpha$ decreases, $S_2$ decreases while $S_2^T$ remains constant, meaning that $S_{12}$ increases. Hence, at low $f$, the interaction of both inputs matters for our optimization problem and not exclusively $X_2$ as found before when considering the whole domain of $X$. The right side of \cref{fig:Sobol_Indic_DixonPr} shows the evolution of the contour of the function which gives an insight of the results obtained previously: while most of the variance in the left plot is due to $X_2$ and the contour lines correspond to those of a function without interaction, the contour lines in the right plot are distorted with a stronger role of $X_1$ and its interaction with $X_2$.

\begin{figure} [h!]
\centering
\includegraphics[width=0.4\linewidth]{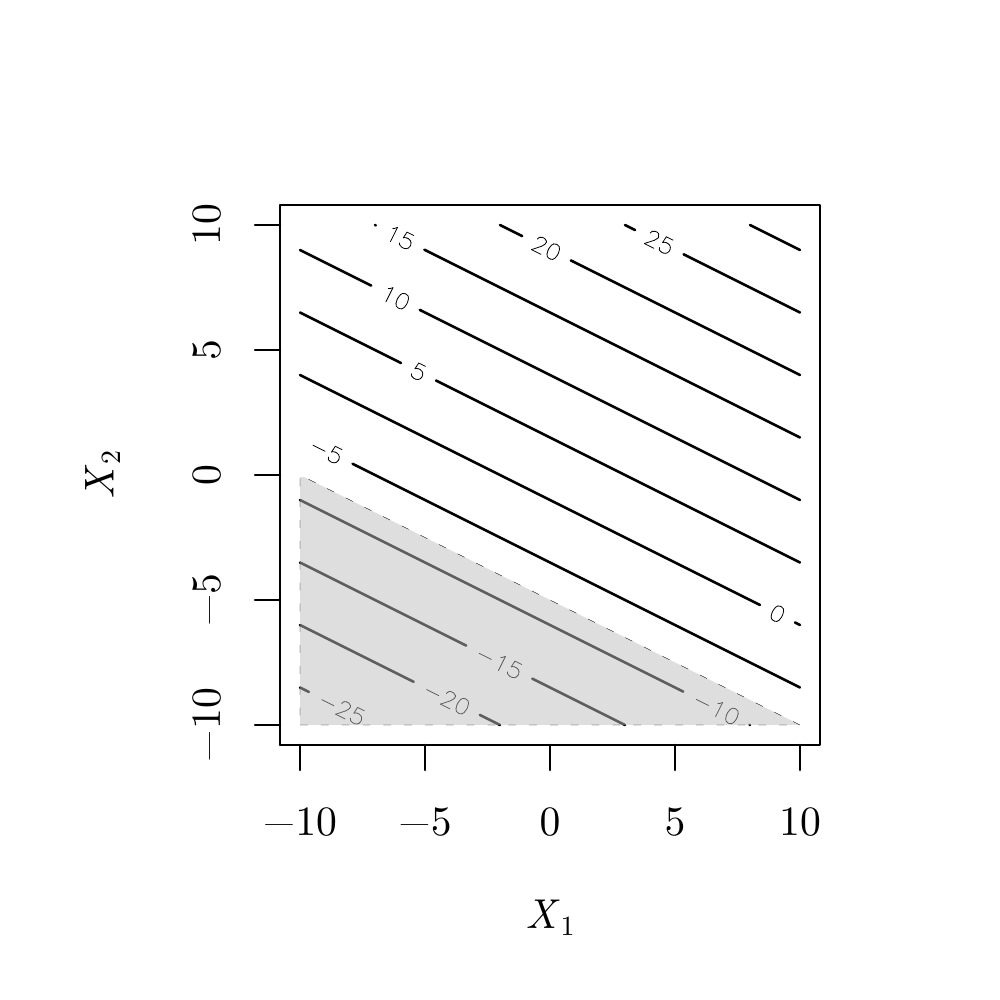}
\caption{Contour plot of the linear function. The gray area shows $\mathcal{D} = \{f(X) \leq q_{25\%}\}$.}
\label{fig:Contour_Linear}
\end{figure}

\begin{figure} [h]
\centering
\includegraphics[width=0.7\textwidth]{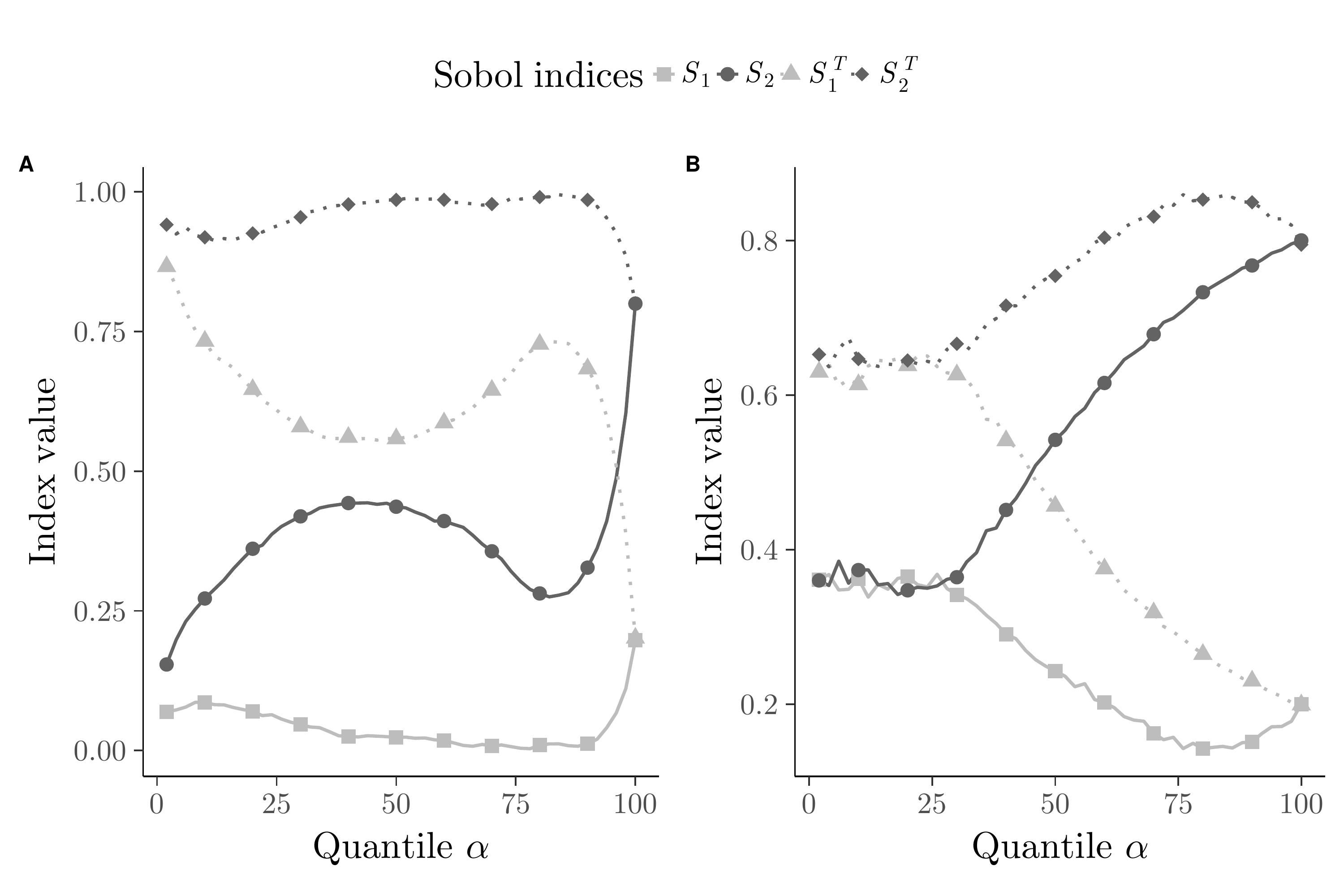}
\caption{Evolution of $S_1$ and $S_1^T$, resp. $S_2$ and $S_2^T$, with respect to the quantile $\alpha$ for the linear function using \textbf{(A)} zero- and \textbf{(B)} conditional-thresholdings.}
\label{fig:Sobol_Indic_Linear}
\end{figure}

Although in the last example the sensitivity of the variable is qualitatively well captured, the zero-thresholding is hard to interpret for two reasons. First of all, the values of $Z$ outside of $\mathcal{D}$ are arbitrarily fixed at zero but other value are possible\footnotemark[1] and this will affect the calculated sensitivities. Second, the sensitivity of $Z$ using this thresholding characterizes both the variation of $f$ inside $\mathcal{D}$ and the shape of $\mathcal{D}$. To illustrate this point let us consider the simple example of a linear function, $f(X) = X_1 + 2X_2$, defined on $[-10, 10]^2$, cf \cref{fig:Contour_Linear}. The sensitivity indices can be analytically determined on the whole domain as $S_1 = 1/5$ and $S_2 = 4/5$. Since the function is already in its decomposed form, it is read that there is no interaction between variables for the complete domain (i.e., $\alpha = 100\%$). Yet, interactions appear when $\alpha$ gets lower than $100\%$, as it can be seen in  \cref{fig:Sobol_Indic_Linear} (A), because $\mathcal{D}$ takes a non-rectangular shape, cf \cref{fig:Contour_Linear}. 

\paragraph{Conditional-thresholding} Unlike the previous zero-thresholding, the formulation of the conditional-thresholding hints that it aims at knowing which inputs are important inside $\mathcal{D}$.  Yet, a dependency on the shape of $\mathcal{D}$ remains, as it can be seen with the linear function of \cref{fig:Sobol_Indic_Linear} (B) where we observe two phenomena. Firstly, for all $\alpha$ below 25\%, the indices reach a steady-state. Below this value of $\alpha$, the shape of $\mathcal{D}$ remains a right-angled triangle of unit high and base length of two, affecting all variables in the same way, see \cref{fig:Contour_Linear}.  Besides, both first order indices are equal from this point on: while $X_2$ is twice more sensitive than $X_1$ in the function definition, its interval in the sub-level $\mathcal{D}$ is twice as narrow, which makes up for the difference in terms of Sobol indices. Prior to $\alpha = 25\%$, the shape of $\mathcal{D}$ depends on $\alpha$. Note that, this version of thresholding does not have	any threshold value, unlike the zero-thresholding where values outside $\mathcal{D}$ were set to zero.

\paragraph{Indicator-thresholding} This last thresholding transformation captures which variables are important in order to reach $\mathcal{D}$ while not depending on the specific values of the objective function $f$ inside it. Since it turns the output into a categorical variable, there is no need to assign a specific value to points outside $\mathcal{D}$. Unlike previous thresholdings, the indicator-thresholding only characterizes the boundary of $\mathcal{D}$ and keeps no information about the values inside or outside the set of interest. It is independent of any monotonous scaling of $f()$, which is a desirable invariance property in optimization \cite{ollivier2011information}. Another advantage of the indicator-thresholding is that one can evaluate the impact of an input through the distance between its distribution and its conditional distribution given $Z = \mathds{1}_{X \in \mathcal{D}}$, which is an approach already taken, although in ways different from the ones we will follow here, in \cite{hornberger1981approach}, and later in \cite{saltelli2004sensitivity}. We further elaborate on that in \cref{ssec:HSIC_Categorical}. From now on, we will only focus on the indicator thresholding because of its aforementioned assets. \\

Whether of not the model outputs are transformed in the above ways,, variance based estimators, because of their formulation (Eq \eqref{eq:Sobol_ind}), can only capture the effect of a variable on the conditional mean of the output. This may fail to reflect the importance of an input in certain cases, especially when the variance is not sufficient to describe the influence of the parameters, e.g. when the output distribution is highly skewed and heavily tailed, see the example in \cite{liu2006relative}.

This motivates us to explore different methods for global sensitivity analysis, methods  based on kernels. These measures are appealing for sensitivity analysis since they weight the impact of a variable on the full distribution of the system output, are well-suited for multi-objective problems and can handle categorical objects with ease. 

\section{Kernel-based global sensitivity analysis}
\label{sec:kbasedSA}
In the following, we introduce the concepts and notation required to understand the kernel-based sensitivity analysis where probability distributions are embedded into reproducing kernel Hilbert spaces (RKHS). Then we define the Maximum Mean Discrepancy (MMD) and the Hilbert-Schmidt independence criterion (HSIC).  

\subsection{Reproducing Kernel Hilbert Space and distribution embeddings}
Let $\mathcal{X}$ be any topological space where a Borel measure can be defined and $\mathcal{H}$ a Hilbert space of $\mathbb{R}$-valued functions on $\mathcal{X}$. $\mathcal{H}$ is called a reproducing kernel Hilbert Space (RKHS), endowed with a dot product $\langle \cdot$, $\cdot \rangle_\mathcal{H}$, if there exists a \emph{kernel} function $k : \mathcal{X} \times \mathcal{X} \to \mathbb{R}$ with the following properties:

\begin{enumerate}
\item $\forall x \in \mathcal{X}$, $k(x,\cdot) \in \mathcal{H}$
\item Reproducing property: $\forall x \in \mathcal{X}$, $\forall f \in \mathcal{H}$ $\langle f, k(x,\cdot) \rangle_\mathcal{H} = f(x)$
\end{enumerate}

As stated in \cite{aronszajn1950theory}, for any symmetric, positive definite function $k : \mathcal{X} \times \mathcal{X} \to \mathbb{R}$, there is a unique associated RKHS, \cite{berlinet2011reproducing}. Furthermore, if one consider a mapping $\phi: \mathcal{X}$ to $\mathcal{H}$ from each $ x \in \mathcal{X}$ to the RKHS $\mathcal{H}$, $k$ can then be expressed as the dot product in term of the feature map, such that, $k(x,x') = \langle \phi(x), \phi(x') \rangle_\mathcal{H}$. Popular reproducing kernels in $\mathbb{R}^d$ include the linear kernel $k(x,x') = \langle x,x' \rangle_\mathcal{X}$, the polynomial kernel of degree $p \in \mathbb{N}$ $k(x,x') = (1+ \langle x,x' \rangle_\mathcal{X})^p$ or the Gaussian radial basis function kernel with bandwidth $\sigma > 0$, $k(x,x') = \exp \left( -\frac{\| x-x'\|^2}{2\sigma^2} \right)$.

We now consider a random variable $X \in \mathcal{X}$ with probability distribution $\mathbb{P}_X$. Following \cite{smola2007hilbert}, we define the kernel embedding $\mu_{\mathbb{P}_X} \in \mathcal{H}$ of $\mathbb{P}_X$ by, 
\begin{equation}
\mu_{\mathbb{P}_X} := \mathbb{E}_X[k(X,\cdot)]= \mathbb{E}_X[\phi(X)],
\end{equation}
provided $\mathbb{E}_X[k(X,X)] < \infty$.

By definition, $\mu_{\mathbb{P}_X}$ is the representation of the distribution $\mathbb{P}_X$ in the RKHS $\mathcal{H}$. Let $\mathcal{M}^1_+(\mathcal{X})$ be the family of all distributions $\mathbb{P}$ on $\mathcal{X}$, the kernel $k$ is called characteristic if the map $\mathcal{M}_k : \mathcal{M}^1_+(\mathcal{X}) \to \mathcal{H}, ~ \mathbb{P}_X \mapsto \mu_{\mathbb{P}_X} = \mathbb{E}_X[k(X,\cdot)]$ is injective \cite{fukumizu2007kernel}. 
Hence, if a characteristic kernel is used, each distribution can be uniquely represented in $\mathcal{H}$ by the kernel embedding and all its statistical features are preserved. 

In the rest of the paper, $\mathbb{P}_X$ will be indirectly specified through its i.i.d. samples $\{ x^1, \ldots, x^N\}$. One must use a Monte-Carlo estimator of the kernel embedding:
\begin{equation}
\hat{\mu}_{\mathbb{P}_X} = \frac{1}{N} \sum_{i=1}^N k(x^i, \cdot)
\end{equation}

\begin{figure}[h!]
\centering
\includegraphics[width=0.8\textwidth]{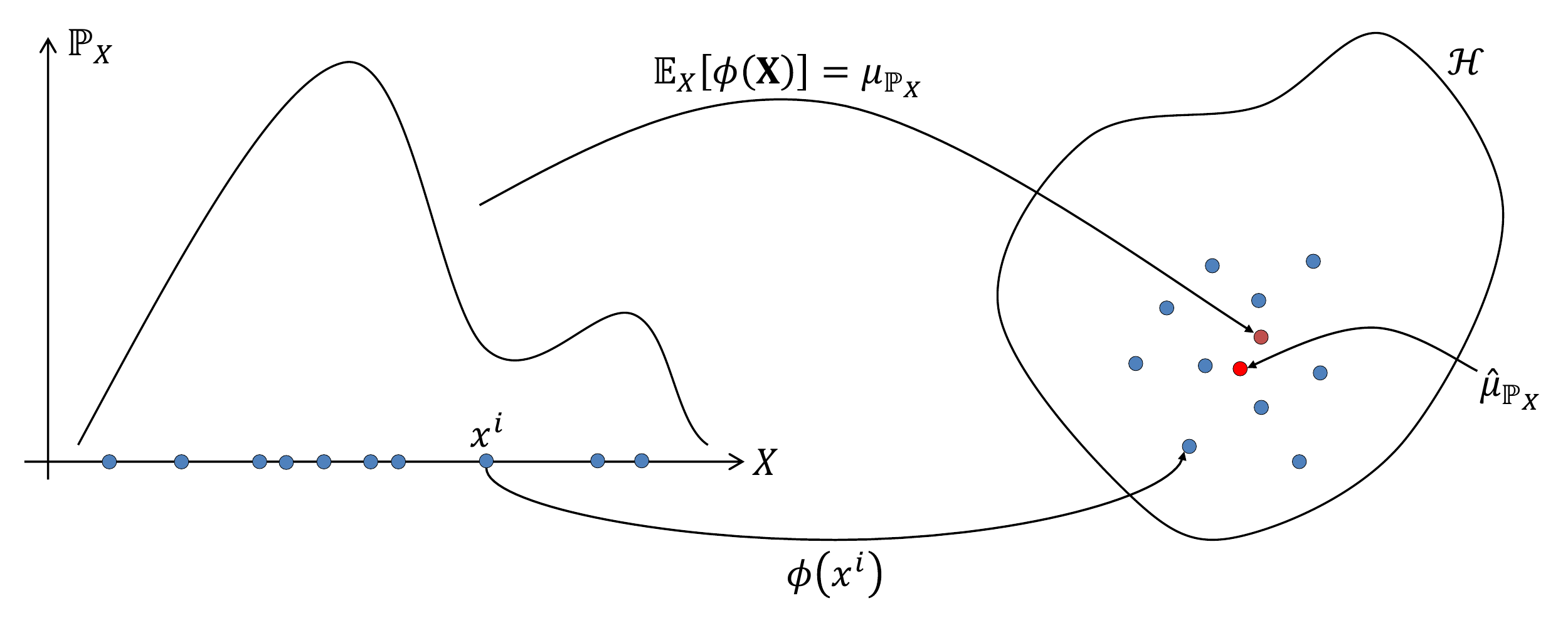}
\caption{Kernel embedding of a distribution, from \cite{song2013kernel}.}
\label{fig:DistributiontoRKHS} 
\end{figure}

One can easily see that $\hat{\mu}_{\mathbb{P}_X}$ converges to $\mu_{\mathbb{P}_X}$ as $N \to \infty$. \cref{fig:DistributiontoRKHS} shows a visual of the kernel embedding of a distribution and its empirical estimator.

\subsection{From Maximum Mean Discrepancy to Hilbert Schmidt Independence}
Kernel embeddings of probability measures provide a distance between distributions as the distance between their embeddings in the Hilbert space. Such distance is called the Maximum Mean Discrepancy, or MMD \cite{gretton2012kernel}. 
Assuming $X$ and $Y$ are two random vectors defined in $\mathcal{X}$ with probabilities distributions $\mathbb{P}_X$ and $\mathbb{P}_Y$, respectively, and $\mathcal{H}$ a RKHS with kernel $k$, \cite{gretton2012kernel} defines the Maximum Mean Discrepancy $\gamma_k$ as the distance between the distribution kernel embeddings in the RKHS by:
\begin{equation}
\gamma_k(\mathbb{P}_X, \mathbb{P}_Y) = \| \mu_{\mathbb{P}_X} - \mu_{\mathbb{P}_Y} \|_{\mathcal{H}}
\end{equation}
In order for $\gamma_k$ to be a metric, i.e. $\gamma_k = 0$ iff $\mathbb{P}_X = \mathbb{P}_Y$, $k$ must be characteristic. This criterion is satisfied by many common kernels (such as the Gaussian kernel, see \cite{sriperumbudur2008injective}, \cite{sriperumbudur2010hilbert} and \cite{fukumizu2009characteristic} for more details). It is common to take the square of the previous equation as it can be expressed in terms of products of expectations. 
\begin{equation}
\gamma^2_k(\mathbb{P}_X, \mathbb{P}_Y) = \mathbb{E}_{X}\mathbb{E}_{X'}[k(X,X')] + \mathbb{E}_{Y}\mathbb{E}_{Y'}[k(Y,Y')] - 2\mathbb{E}_{X}\mathbb{E}_{Y}[k(X,Y)]
\end{equation}
for $X'$ and $Y'$ independent copies of $X$ and $Y$, s.t. $X,X' \sim \mathbb{P}_X$ and $Y,Y' \sim \mathbb{P}_Y$. The previous equation can also be written under an integral form, often useful for proofs, as
\begin{equation} \label{IntegralMMD}
\begin{aligned}
\gamma^2_k(\mathbb{P}_X, \mathbb{P}_Y) & = \iint_\mathcal{X} k(x,y) d(\mathbb{P}_X - \mathbb{P}_Y)(x)(\mathbb{P}_X - \mathbb{P}_Y)(y) \\
									   & = \iint_\mathcal{X} k(x,y) (p_X(x) - p_Y(x))(p_X(y) - p_Y(y))dxdy
\end{aligned}
\end{equation}

Furthermore, the MMD can be used to detect statistical dependencies between random variables (\cite{gretton2005measuring}, \cite{gretton2007kernel} or \cite{smola2007hilbert} for example). Assume $\mathcal{H}$ is a RKHS of functions from $\mathcal{X}$ to $\mathbb{R}$ with kernel $k$. Likewise, we define a second RKHS, $\mathcal{G}$, of functions from $\mathcal{Y}$ to $\mathbb{R}$ with kernel $l$. Because of product of kernels is a kernel \cite{gpml2006}, we construct the kernel $v$ on the product space $\mathcal{X} \times \mathcal{Y}$ with RKHS $\mathcal{V}$ as:
\begin{equation}
v((x, y),(x',y')) = k(x, x')l(y, y')
\end{equation}

Let $X \sim \mathbb{P}_X$ be a random variable on $\mathcal{X}$ and a RKHS $\mathcal{H} : \mathcal{X} \to \mathbb{R}$ induced by a kernel $k$. Similarly, let $Y \sim \mathbb{P}_Y$ be a random variable on $\mathcal{Y}$ and a RKHS $\mathcal{G} : \mathcal{Y} \to \mathbb{R}$ induced by a kernel $l$. $X$ and $Y$ have a joint distribution $\mathbb{P}_{XY}$. Given $v$ a kernel on $\mathcal{X} \times \mathcal{Y}$ as defined above, the Hilbert-Schmidt Independence Criterion (HSIC) of $X$ and $Y$ is the squared MMD $\gamma_v^2$ between $\mathbb{P}_{XY}$ and the product of its marginals $\mathbb{P}_X\mathbb{P}_Y$. Following \cite{smola2007hilbert}, we begin by defining the distribution embeddings of $\mathbb{P}_{XY}$ and $\mathbb{P}_X\mathbb{P}_Y$:
\begin{equation}
\begin{aligned}
\mu_{\mathbb{P}_{XY}} & = \mathbb{E}_{XY}[v((X,Y), \cdot)] \\
\mu_{\mathbb{P}_{X}\mathbb{P}_{Y}}	& = \mathbb{E}_{X}\mathbb{E}_{Y}[v((X,Y), \cdot)]
\end{aligned}
\end{equation}
From this, we write HSIC as 
\begin{equation}
\begin{aligned}
\textrm{HSIC}(X,Y)_{\mathcal{H},\mathcal{G}} = &  ~ \gamma_v^2(\mathbb{P}_{XY},\mathbb{P}_X\mathbb{P}_Y) = \| \mu_{\mathbb{P}_{XY}} -  \mu_{\mathbb{P}_{X}\mathbb{P}_{Y}}\|_{\mathcal{H} \otimes \mathcal{G}}^2 \\
 = & ~ \mathbb{E}_{X,Y} \mathbb{E}_{X',Y'} k(X,X')l(Y,Y') \\
 & + \mathbb{E}_{X} \mathbb{E}_{X'} \mathbb{E}_{Y} \mathbb{E}_{Y'} k(X,X')l(Y,Y') \\
 & - 2 \mathbb{E}_{X,Y} \mathbb{E}_{X'} \mathbb{E}_{Y'} k(X,X')l(Y,Y')
\end{aligned}
\end{equation}
The latter is also the squared Hilbert-Schmidt norm of the cross-covariance operator associated with $\mathbb{P}_{XY}$ between RKHSs \cite{gretton2005measuring}. For characteristic kernels, $\textrm{HSIC}(X,Y)_{\mathcal{H},\mathcal{G}}$ is zero if and only if X and Y are independent.

Like the MMD, the HSIC can be written as an integral
\begin{equation} \label{eq:HSICIntegral}
\begin{aligned}
\textrm{HSIC}(X,Y)_{\mathcal{H},\mathcal{G}} 
	= & \int_\mathcal{X} \int_\mathcal{Y} k(x,x') l(y,y') d(\mathbb{P}_{XY} - \mathbb{P}_X\mathbb{P}_Y)(x,y)(\mathbb{P}_{XY} - \mathbb{P}_X\mathbb{P}_Y)(x,y') \\
	= & \int_\mathcal{X} \int_\mathcal{Y} k(x,x') l(y,y') (p_{XY}(x,y) - p_X(x)p_Y(y)) \\
	& \times (p_{XY}(x',y') - p_X(x')p_Y(y')) dxdx'dydy'
\end{aligned}
\end{equation}

\cite{gretton2005measuring} also proposed a HSIC empirical estimator for a finite number $N$ of observations. Let $(\textbf{X},\textbf{Y}) = \{ (x^1, y^1), \ldots, (x^N, y^N) \} \subseteq \mathcal{X} \times \mathcal{Y}$ be independently drawn samples from $\mathbb{P}_{XY}$, we have $\textrm{HSIC}(\textbf{X},\textbf{Y})_{\mathcal{H},\mathcal{G}}$ as
\begin{equation} \label{HSICestimator}
\textrm{HSIC}_N(\textbf{X},\textbf{Y})_{\mathcal{H},\mathcal{G}} = \frac{1}{N^2}\textrm{tr}(KHLH) 
\end{equation}
where $K$, $L \in \mathbb{R}^{N \times N}$, $K_{ij} = k(x^i, x^j)$, $L_{ij} = l(y^i, y^j)$ are kernel matrices and $H \in \mathbb{R}^{N \times N}$, $H = I_N - N^{-1}\textbf{11}^T$ is a centering matrix. \textbf{1} is a vector of ones of proper dimension. The proposed estimator is computed in $O(N^2)$ time.

In the global sensibility analysis framework, \cite{da2015global} proposes a sensitivity index based on a normalized version of the HSIC measure between $f(X)$ and $X_i$, which is further studied in \cite{de2014new}.

\subsection{HSIC sensitivity index with categorical inputs} \label{ssec:HSIC_Categorical}

The above HSIC sensitivity index relies only on the choice of the kernel functions for the inputs and the outputs. This choice depends directly on the type of data: for example, for continuous data sets, it is customary to use the squared exponential kernel,
\begin{equation*}
k(x,x') = \exp \left( -\frac{\| x - x'\|_2^2}{2\sigma^2}\right)
\end{equation*}
with $\sigma$ the bandwidth parameter. 

Applying the indicator-thresholding introduced in \cref{ssec:formulation} to the output $Z = \mathds{1}_{f(X) \leq q \cap \textbf{g}(X) \leq \textbf{T}}$ makes it a binary variable. Kernels dedicated to categorical variables, such as the linear kernel $l(z,z') = zz'$, can then be employed. 
In that case, one can draw a link between the HSIC and a MMD. With $Z = \mathds{1}_{X \in \mathcal{D}}$, the HSIC expression of \eqref{eq:HSICIntegral} becomes
\begin{equation} \label{eq:HSIC_Categorical}
\begin{aligned}
\text{HSIC}(X,Z)
	= & \int_{x,x'} \sum_{z=0}^{1} \sum_{z'=0}^{1} k(x,x')l(z,z') \left[p_{X|Z=z}(x) - p_{X}(x)\right] \\
	   & \times \left[p_{X|Z=z'}(x') - p_{X}(x')\right] \mathbb{P}{(Z=z)}\mathbb{P}{(Z=z')}dxdx'
\end{aligned}
\end{equation}

We can simplify the previous equation thanks to zero values of $l(z,z')$ and obtain
\begin{equation}
\begin{aligned}
\text{HSIC}(X,Z) = &\int_{x,x'} k(x,x') \left[p_{X|Z=1}(x) - p_{X}(x)\right] \\
			& \times \left[p_{X|Z=1}(x') - p_{X}(x')\right] \mathbb{P}(Z=1)^2 dxdx'
\end{aligned}
\end{equation}
Recalling the integral expression of the Maximum Mean Discrepancy \eqref{IntegralMMD}, one notices that
\begin{equation}
\text{HSIC}(X,Z) = \mathbb{P}(Z=1)^2 \times \gamma_{k}^2(\mathbb{P}_{X|Z=1},\mathbb{P}_X)
\label{eq:HSICtoMMD}
\end{equation}
Eq \eqref{eq:HSICtoMMD} shows the equivalence between the distance in distribution of $X$ and $X | Z = 1$ and the statistical dependence of $X$ and $Z$. \eqref{eq:HSICtoMMD} is similar to the concepts behind the Monte Carlo Filtering technique for sensitivity analysis first introduced in \cite{hornberger1981approach}. 

In order to evaluate the importance of each variable separately, the input used in \eqref{eq:HSICtoMMD} is $X_i$, which directly defines a sensitivity measure from HSIC and the indicator thresholding. 

\begin{definition}[Sensitivity index from HSIC with Indicator-Thresholding, HSIC-IT]
~\\ Let $f(): \mathbb{R}^d \rightarrow \mathbb{R}$ and $g(): \mathbb{R}^d \rightarrow \mathbb{R}^{m} $ be objective and constraints functions of the random variables $X = (X_1, \dots , X_d)$. The sensitivity index of the variable $X_i$ from the Hilbert-Schmidt Independence Criterion with the Indicator-Thresholding (HSIC-IT) is defined as 
\begin{equation}
\Shsici(X_i) = \text{HSIC}(X_i,\mathds{1}_{f(X) \leq q \cap \textbf{g}(X) \leq \textbf{T}})
\end{equation}
for any $q \in \mathbb{R}$ and $\textbf{T} \in \mathbb{R}^{m,+}$. 
\end{definition}
\vskip\baselineskip
\Shsici measures the impact of an input through how much its probability distribution changes when it is restricted by the output. A variable is negligible for optimization if its \Shsici is close to zero. The main difference between the above \Shsici definition and the sensitivity index proposed in \cite{da2015global} lies in the use of the indicator function. In the numerical applications provided in this paper, each $\Shsici(X_i)$ is normalized by dividing with $\sum_{i=1}^d \Shsici(X_i)$. Alternatively, like in \cite{da2015global}, $\Shsici(X_i)$ could be divided by $\sqrt{\text{HSIC}(X_i,X_i) \text{HSIC}(Z,Z)}$.
\begin{figure} [h]
\centering
   \begin{minipage}{.5\linewidth}
   	\centering
   	\vspace{-1.3cm} 
   	\includegraphics[width=\linewidth]{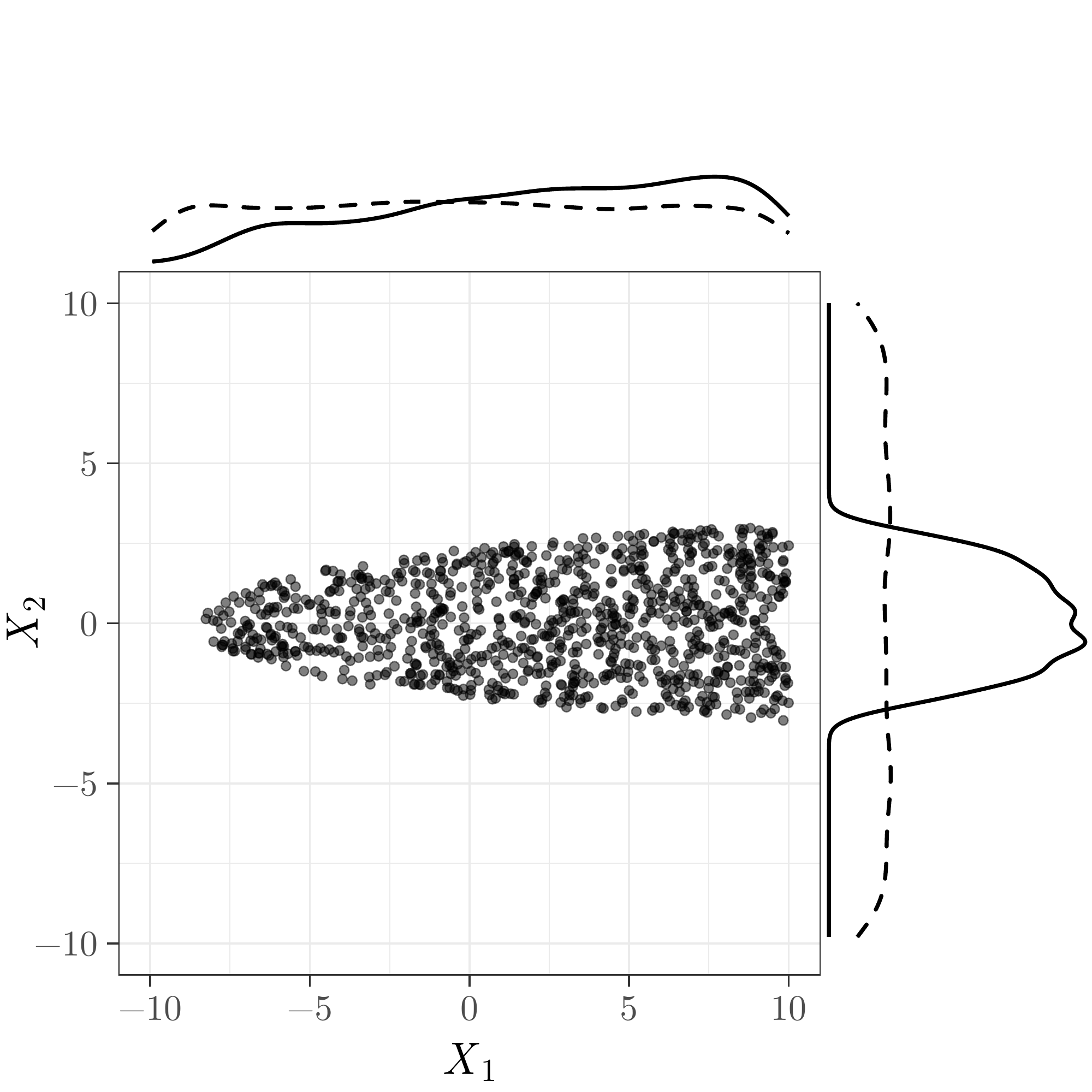}
   \end{minipage}%
   \begin{minipage}{.5\linewidth}
   	\centering
   	\includegraphics[width=0.8\linewidth]{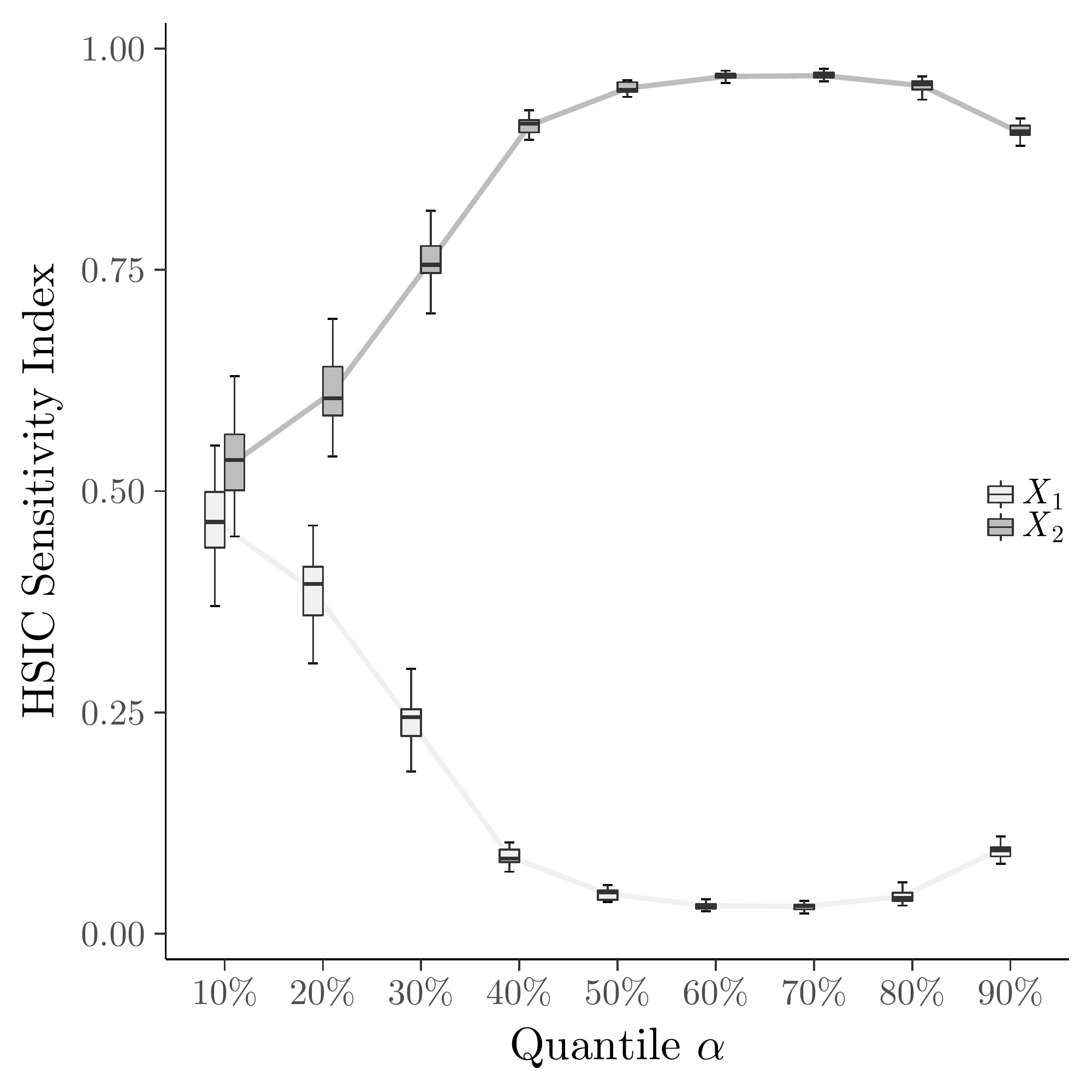}
   \end{minipage}
\caption{\textbf{Left:} Samples from $\mathbb{P}_{f(X) | f(X) \leq q_{\alpha = 20\%}}$ and associated inputs marginal distributions for the Dixon-Price function. The original empirical distribution on the complete domain is also drawn in dashed lines. It is not completely uniform because of the finite size of the sample. \textbf{Right:} Evolution of the normalized HSIC sensitivity indices w.r.t. the quantile $\alpha$.}
\label{fig:DixonPrwithMarginals}
\end{figure}

As an example, consider the two-dimensional Dixon-Price function already discussed in \cref{fig:Sobol_Indic_DixonPr}. \cref{fig:DixonPrwithMarginals} shows the distribution of $\mathbb{P}_{X_1 | Z = 1}$ and $\mathbb{P}_{X_2 | Z = 1}$ for $Z = \mathds{1}_{f(X) \leq q_{\alpha = 20\%} }$ along with the initial inputs distribution, $X_i \sim \mathcal{U}[-10,10]$, for $i = \{1,2\}$. Because the distribution of $\mathbb{P}_{X_2 | Z = 1}$ differs more from the uniform one than that of $\mathbb{P}_{X_1 | Z = 1}$, its sensitivity is larger. 

Another illustration is given by the following two-dimensional ``Level'' function whose behavior changes at a certain threshold $q$: above the threshold $q$, $f(X)$ only depends on $X_1$ but it only depends on $X_2$ below $q$,
\begin{equation*}
f(X) = 
 \begin{cases}
	\lvert X_1 \rvert  &\quad \text{if} \quad \lvert X_1 \rvert > q \\
	\lvert X_2 - 2 \rvert - 6  &\quad \text{otherwise}
 \end{cases}
\end{equation*}

\begin{figure}[h]
\centering
   \begin{minipage}{.5\linewidth}
   	\centering
   	\includegraphics[width=0.8\linewidth]{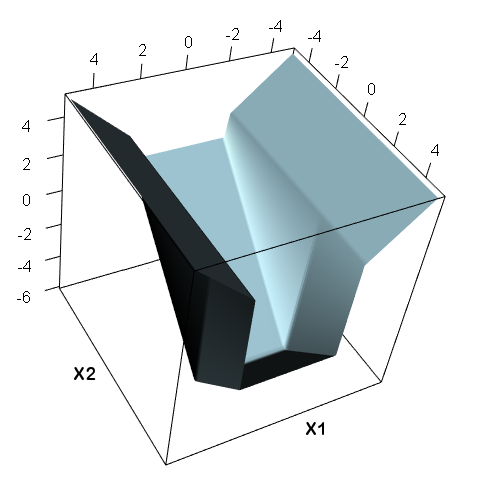}
   \end{minipage}%
   \begin{minipage}{.5\linewidth}
   	\centering
   	\includegraphics[width=0.8\linewidth]{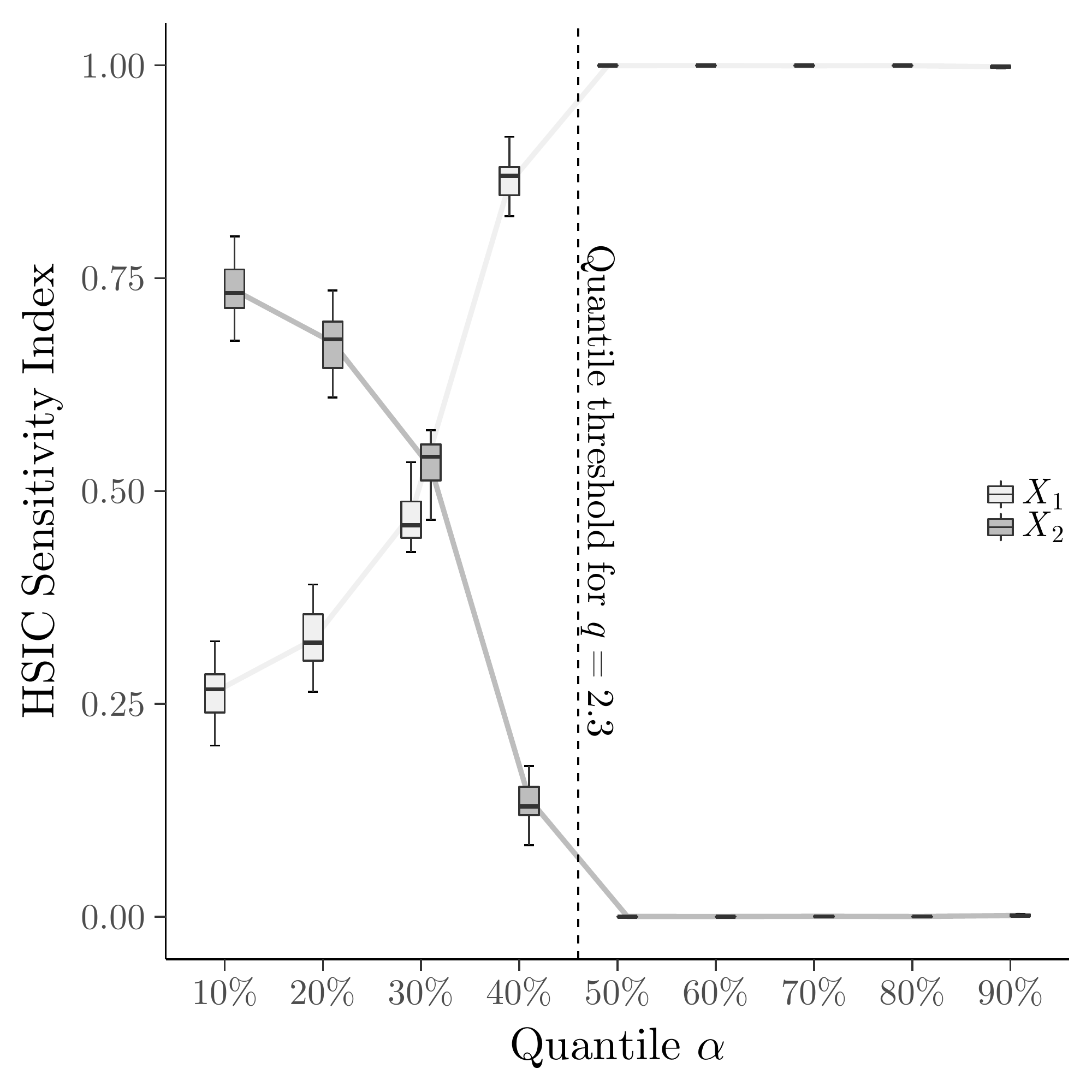}
   \end{minipage}%
\caption{\textbf{Left:} Level-set function for $q = 2.3$. \textbf{Right:} Evolution of the normalized HSIC sensitivity indices for the level-set function w.r.t. the quantile $\alpha$.}
\label{fig:HSIC_level_set}
\end{figure}

The left side of \cref{fig:HSIC_level_set} shows a representation of the ``Level'' function with $q=2.3$ on $\mathcal{S} = [-5,5]^2$. The right side of \cref{fig:HSIC_level_set} provides the sensitivities $\Shsici[q_\alpha](X_i)$, $i = 1, \ldots, 2$, for multiple $\alpha$-quantile values. The dotted line corresponds to the threshold of 2.3. It is observed that the unique dependency on $X_1$ is captured above that threshold where the sensitivity on $X_2$ is null, while both variables have a non-zero sensitivity below the threshold. Indeed, $X_2$ is negligible for reaching the set above $q$. Yet, below $q$, both inputs matter since $X_1$ is necessary to attain that area in a first place and $X_2$ matters to reach sub-areas below $q$.

Hence, the sensitivity measure from HSIC with indicator, \Shsici, detects which inputs are important to arrive at a specific sublevel set.

\section{From HSIC to optimization, with applications}
\label{sec:examples}

\subsection{An optimization strategy based on the HSIC-IT sensitivities}

The HSIC-IT measures naturally lead to an optimization strategy: the HSIC-IT are first calculated and, second, used to simplify the optimization problem, as explained below.
\paragraph*{Sensitivity analysis}

As a preliminary step to the optimization, a sensitivity analysis is done in order to measure which inputs actually matter to reach a certain level of the objective function within the feasible space. We create at random $N$ points making the initial design of experiments $\textbf{X} \in \mathbb{R}^{N \times d}$ and compute the $\Shsici[q_\alpha,T](X_i)$, $i = 1, \ldots, d$, for multiple $\alpha$ values (typically $\alpha = [10\%, 40\%$, $70\%, 100\%]$). The value of $\textbf{T}$ is chosen to ensure at least one hundred feasible points. A Gaussian radial basis function (RBF) kernel is used for the inputs and a linear kernel for the output, $Z = \mathds{1}_{f(x) \leq q_\alpha \cap \textbf{g}(x) \leq \textbf{T}}$. The bandwidth parameter $\sigma$ of the Gaussian RBF kernel is chosen as the median distance between points in the sample set $X$ since it performs well in many applications \cite{gretton2012kernel}. 

\paragraph*{Modification of the optimization problem}

After the sensitivity analysis, an input $X_j$ is dubbed ``negligible'' for the optimization when its normalized $\Shsici[q_\alpha,\textbf{T}](X_j)$ is below a threshold $\tau = 0.1 \times \max_{i=1,\ldots,d} \Shsici[q_\alpha,\textbf{T}](X_i)$ for a low $\alpha$ (here $\alpha = 10 \%$). We set those inputs to a chosen value and reformulate the optimization problem: let \Aset be the index set of active optimization variables whose \Shsici is above $\tau$, and \Acomp the complementary set of fixed variables, so that the initial number of variables is split into $d = \card(\Aset) + \card(\Acomp)$. The modified optimization problem is
\begin{equation}
\begin{aligned}
& \underset{X_i,~ i \in \Aset}{\text{minimize}} & & f(X) \\
& \text{where } & & X_j = x_j ~\text{ is given, } j \in \Acomp \text{ , } X \in \mathcal X ~,\\
& \text{subject to} & & g_l(X) \leq 0, \; l = 1, \ldots, m.
\end{aligned}
\label{eq-opt_modified}
\end{equation}
Two approaches for setting the non-active variables are studied:
\begin{itemize}
\item \emph{Random strategy}: the negligible inputs, $x_j,~j \in \Acomp$, are uniformly sampled from the restriction of $\mathcal X$ to its $j$th component once at the beginning of the search.
\item \emph{Greedy strategy}: the negligible inputs are set to the values provided by the best feasible point of the sensitivity analysis; $x_j,~j \in \Acomp$ is the $j$-th component of \\ $\arg \min_{x^i, i=1,N \mid \vec{g}(x^i)\le 0} f(x^i)$. 
\end{itemize}

The optimization is carried out with the COBYLA algorithm. COBYLA is a local derivative-free algorithm for constrained problem which employs linear approximations to the objective and constraint functions \cite{powell1994direct}. The implementation from the \pkg{nlopt} package \cite{ypma2014introduction} of the \textsf{R} language was used.

\subsection{Constrained optimization test problems}
\subsubsection{Optimization study protocol} 
Tests will be carried out to compare the \emph{Random} and \emph{Greedy} problem formulations, to which we add the unmodified version of the problem, referred to as \emph{Original}, where all $d$ variables are optimized. Because of the small size of the feasible space in the examples below, a large number of points is required for the sensitivity analysis: results are provided for $N = 50000$ and each estimation of $\Shsici[q_\alpha,\textbf{T}]$ is repeated 20 times to obtain a confidence interval on each index. Then, for each optimization problem version (\emph{Original}, \emph{Random} and \emph{Greedy}), the algorithm is run 10000 times. The starting points are each of the 100 points of an optimized Latin Hypercube Sampling (LHS) with maximin criterion\footnotemark[2], with 100 repetitions. 
\footnotetext[2]{Note that it would not be possible to use the same design of experiments for all problem versions since it cannot be guaranteed that the filling criterion is respected after simplification.}
The comparisons will be based on the number of calls to the objective function at convergence and the performance of the solutions.

The following examples are two well-known engineering design test problem: the Gas Transmission Compressor Design Problem \cite{charles1976applied} and the Welded Beam Design Problem \cite{deb2000efficient}. \cref{tab:RecapFeatures} summarizes characteristic features of both problems: the number of inputs $d$, the number of constraints $m$, the ratio in percent of the volume of feasible region to the volume of the complete design space, the best known feasible objective value and the corresponding $X^*$. 

\begin{table}
\caption{Constrained optimization test problems.}
\label{tab:RecapFeatures} 
\centering
\begin{tabular}{|c|c|c|c|c|c|}
\hline 
Name & $d$ & $m$ & \% feas. space & Best $f$ & Best known $X^*$\\ 
\hline 
GTCD & 4 & 1 & $52.38$ & $2964893.85$ & $[49.99, 1.178$,$ 24.59, 0.389]$\\ 
\hline 
WB4 & 4 & 5 & $5.6 \cdot 10^{-2}$ & $1.7250$ & $[0.206, 3.473, 9.037, 0.206]$\\ 
\hline 
\end{tabular}
\end{table}

\subsubsection{Gas Transmission Compressor Design (GTCD)}

The first example is a real-life problem about the design of a gas pipe line transmission system. The objective is to minimize its cost $f(X_1, X_2, X_3, X_4)$ under a nonlinear constraint. The problem objective function, constraint and search space are given below: 
\begin{align*}
f(X) = & (8.61 \times 10^5)X_1^{1/2}X_2X_3^{-2/3}X_4^{-1/2}  + (7.72 \times 10^8)X_1^{-1}X_2^{0.219} \\
				& - (765.43 \times 10^6)X_1^{-1} + (3.69 \times 10^4)X_3
\end{align*}
\noindent s.t. 
\begin{equation*}
g_1(X) = X_4X_2^{-2} + X_2^{-2} - 1 \leq 0
\end{equation*}
\begin{equation*}
20 \leq X_1 \leq 50, ~ 
1 \leq X_2 \leq 10, ~ 
20 \leq X_3 \leq 50, ~ 
0.1 \leq X_4 \leq 60
\end{equation*}

\begin{figure}
\centering
\vspace{1cm}
\begin{tikzpicture}
\node[anchor=south west,inner sep=0] (image) at (0,0) {\includegraphics[width=\textwidth]{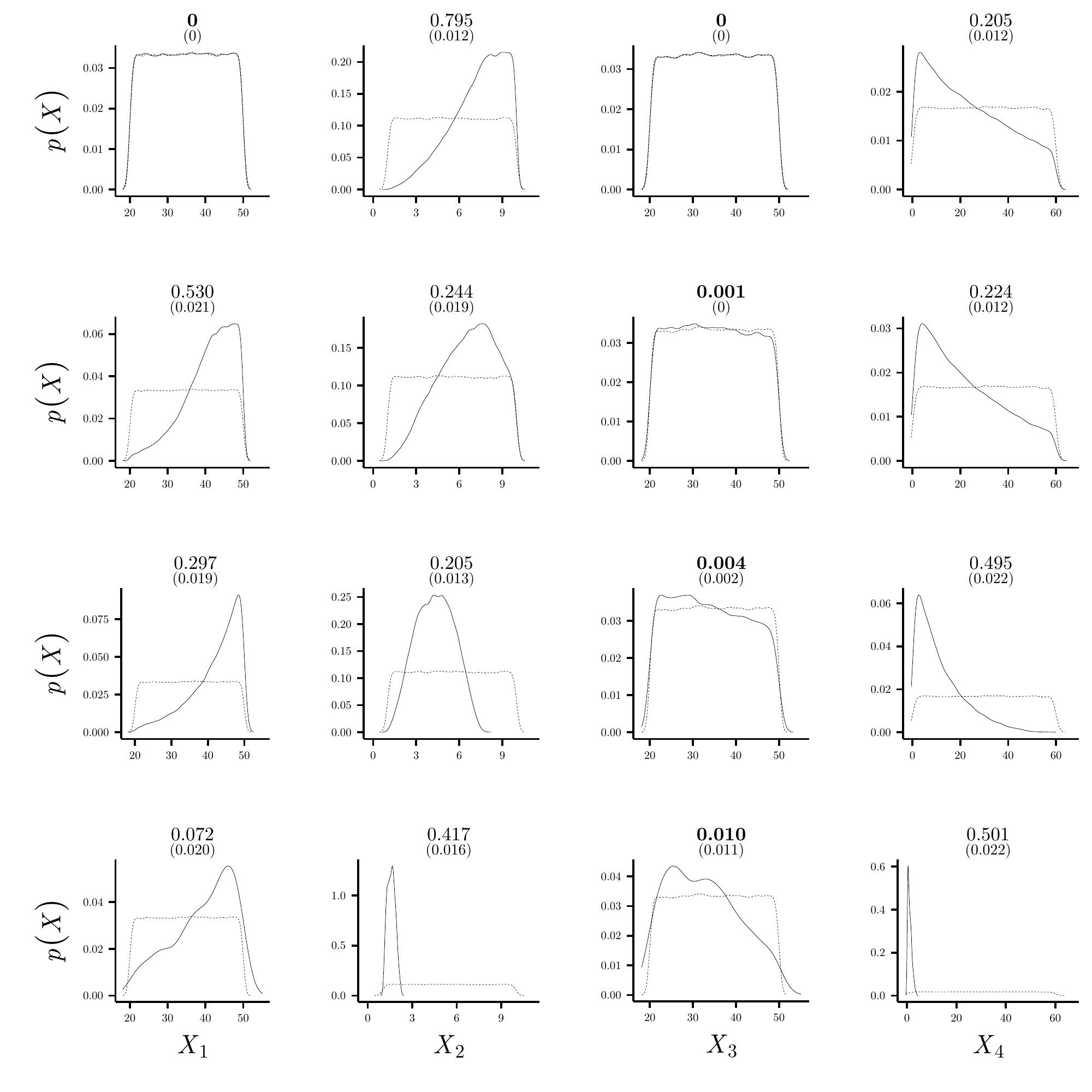}};
    \begin{scope}[x={(image.south east)},y={(image.north west)}]
        \node[inner sep=0pt] at (0.07,0.24){$\alpha = 10\%$};
		\node[inner sep=0pt] at (0.07,0.49){$\alpha = 40\%$};
		\node[inner sep=0pt] at (0.07,0.74){$\alpha = 70\%$};
		\node[inner sep=0pt] at (0.07,0.985){$\alpha = 100\%$};
    \end{scope}
\end{tikzpicture}
\caption{Evolution of $\mathbb{P}_{X_i | Z = 1}$ for different $\alpha$ values (continuous line) compared to the original distribution (dashed line) for the Gas Transmission Compressor Design. The continuous and dashed lines differ for $\alpha = 100\%$ because all points are not feasible. The numbers above each plot are the corresponding mean and standard deviation of $\Shsici[q_\alpha,\textbf{T}]$. Bold numbers correspond to negligible $X_i$'s, i.e., small $\Shsici[q_\alpha,\textbf{T}]$'s. For better readability, the scales of the vertical axes vary.}
\label{fig:Dist_HSIC_GTCD}
\end{figure}
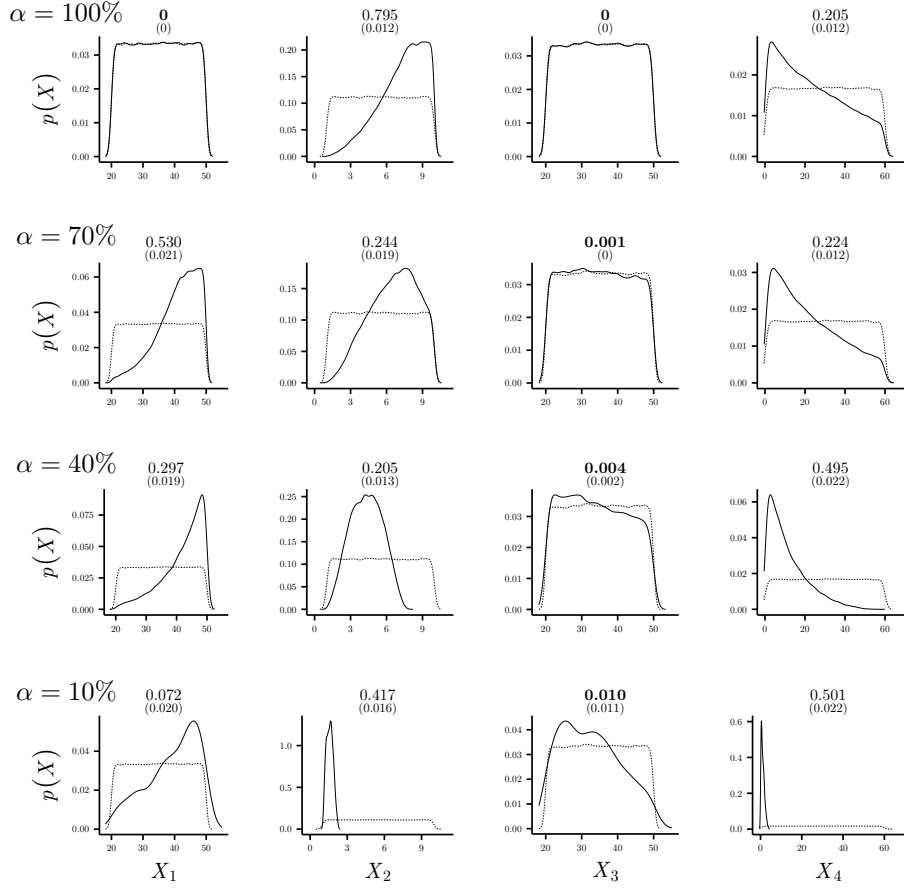

\paragraph{Sensitivity analysis} \cref{fig:Dist_HSIC_GTCD} shows the evolution of the conditional distributions for different quantiles $\alpha$. Above each plot, the corresponding means and standard deviations, out of the 20 repetitions, of the normalized HSIC-IT sensitivities are given. $X_3$ is detected as negligible as its index is near zero for the low quantile $\alpha = 10\%$. The near zero HSIC-IT of $X_3$ expresses the fact that its conditional probability distribution stays relatively close to the uniform distribution while other inputs see their distribution become increasingly skewed as $\alpha$ decreases. 

The value chosen for $X_3$ in the \emph{Greedy} modification of the optimization problem is $29.19$ as it returned the best objective value during the sensitivity analysis, with the best point known in the literature being $X^* = [49.99, 1.178$,$ 24.59, 0.389]$. 

\begin{figure}
\centering
\scalebox{0.8}{
\begin{minipage}[c]{\linewidth}
\centering
\centerline{\includegraphics[scale = 0.6]{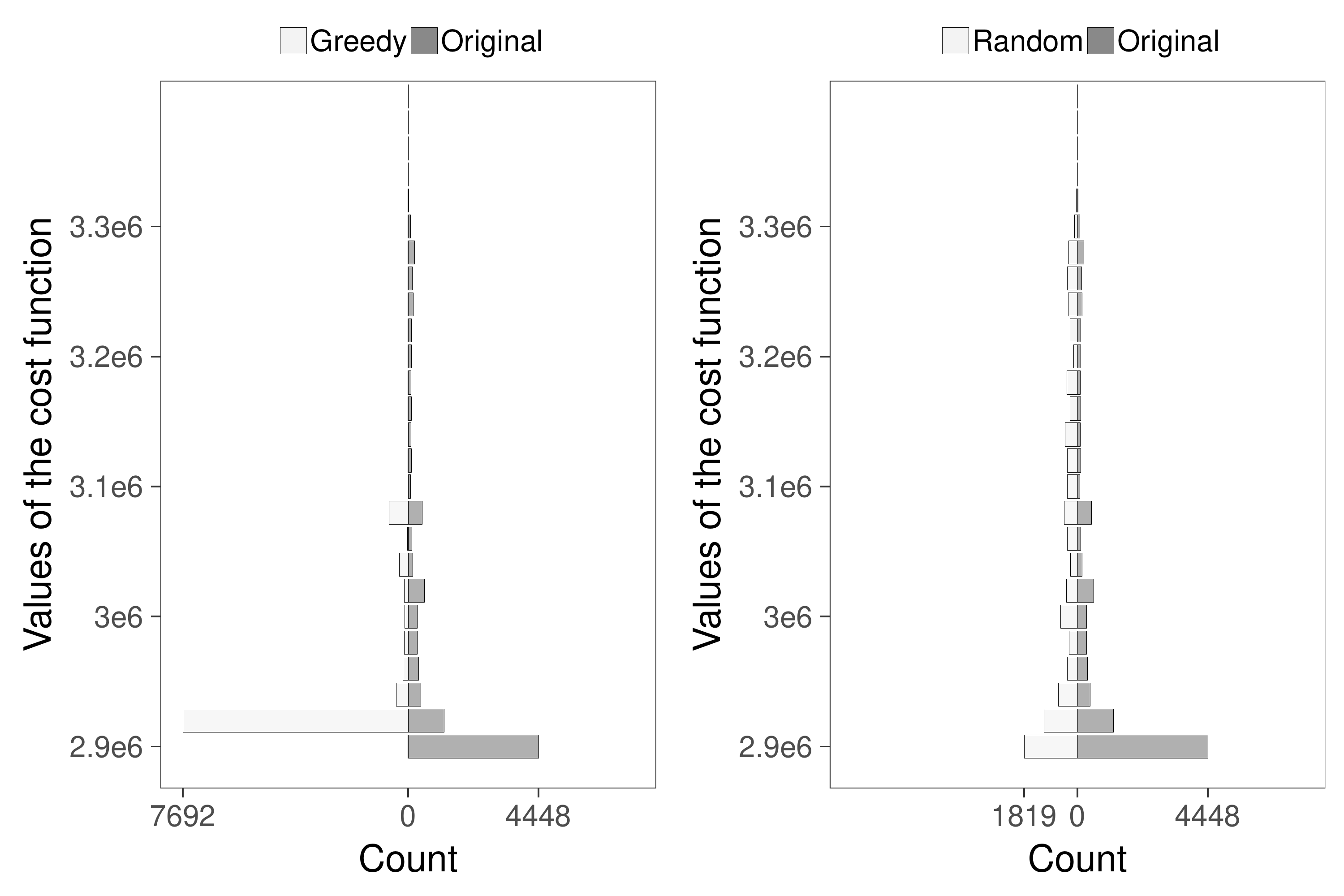}}
\centerline{\includegraphics[scale = 0.6]{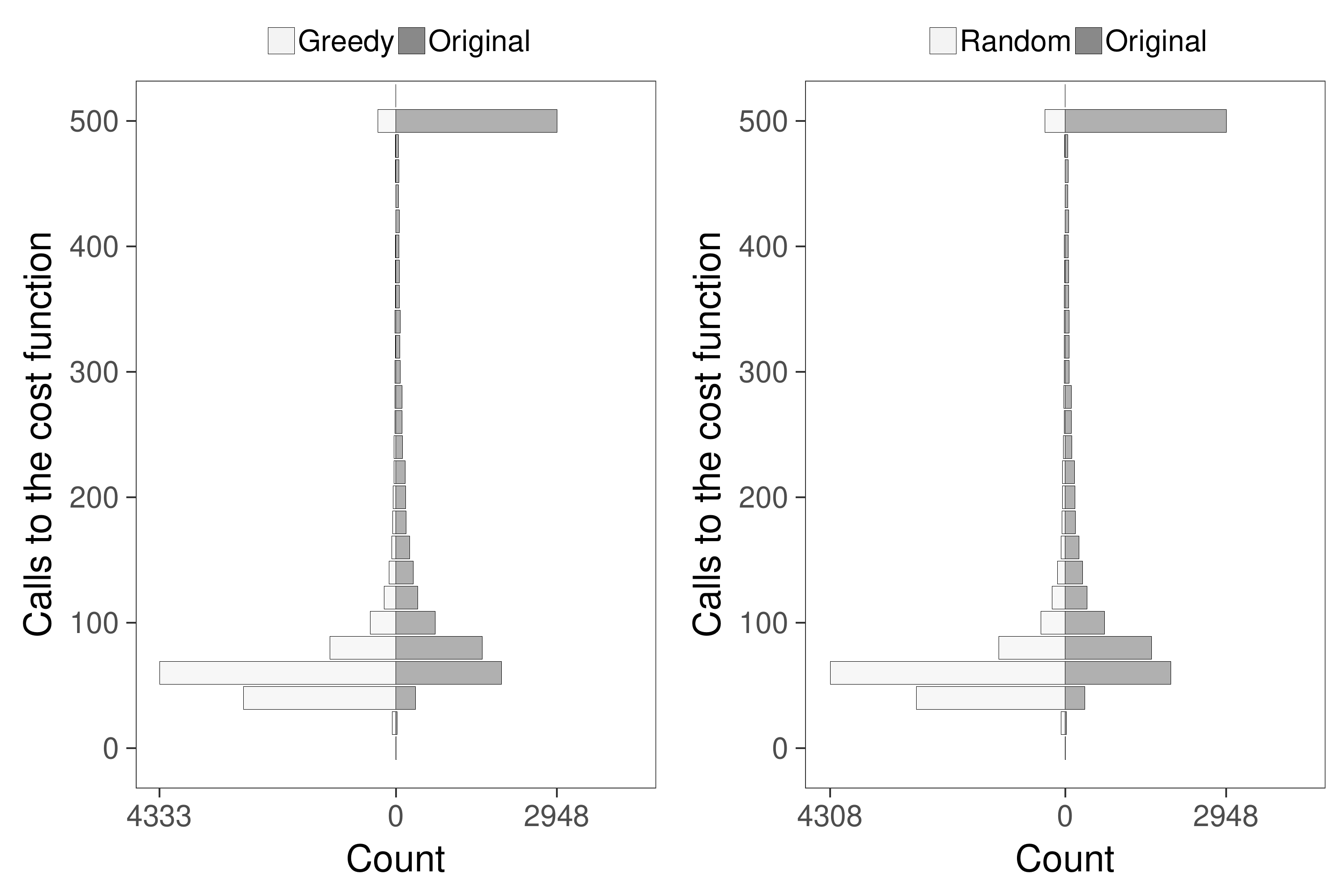}}
\end{minipage}}
\caption{Results of 10000 optimizations with the Original, Greedy and Random formulations for the Gas Turbine Compressor Design test case: histograms of the final objective functions (top) and number of calls to the objective function at convergence (bottom). }
\label{fig:results_optim_GTCD}
\end{figure}

\paragraph{Optimization results} 
\cref{fig:results_optim_GTCD} sums up the results of the 10000 optimization runs. The histograms on the top left plot show that the \emph{Greedy} version of the problem has a degraded optimum, $f(X) = 2966731$, with respect to the \emph{Original} one, $f(X) = 2964894$. But the convergence is more robust, showing fewer runs that get trapped at local solutions and it is obvious from the lower plots that convergence is faster. The \emph{Random} version has a cost similar to that of the \emph{Greedy} formulation, but the cost functions at convergence vary significantly depending on the values chosen at random for the frozen inputs.
Both modified versions of the problem use significantly fewer calls to the objective function than the original formulation as it might be expected from problems with smaller search spaces. Although they return inferior solutions, the \emph{Greedy} formulation is acceptable since it yields solutions close to the original optimum in a faster and more consistant manner.

\subsubsection{Welded Beam (WB4)}

\paragraph{Optimization problem}
\begin{figure} [h!]
\centering
\includegraphics[width=0.6\textwidth]{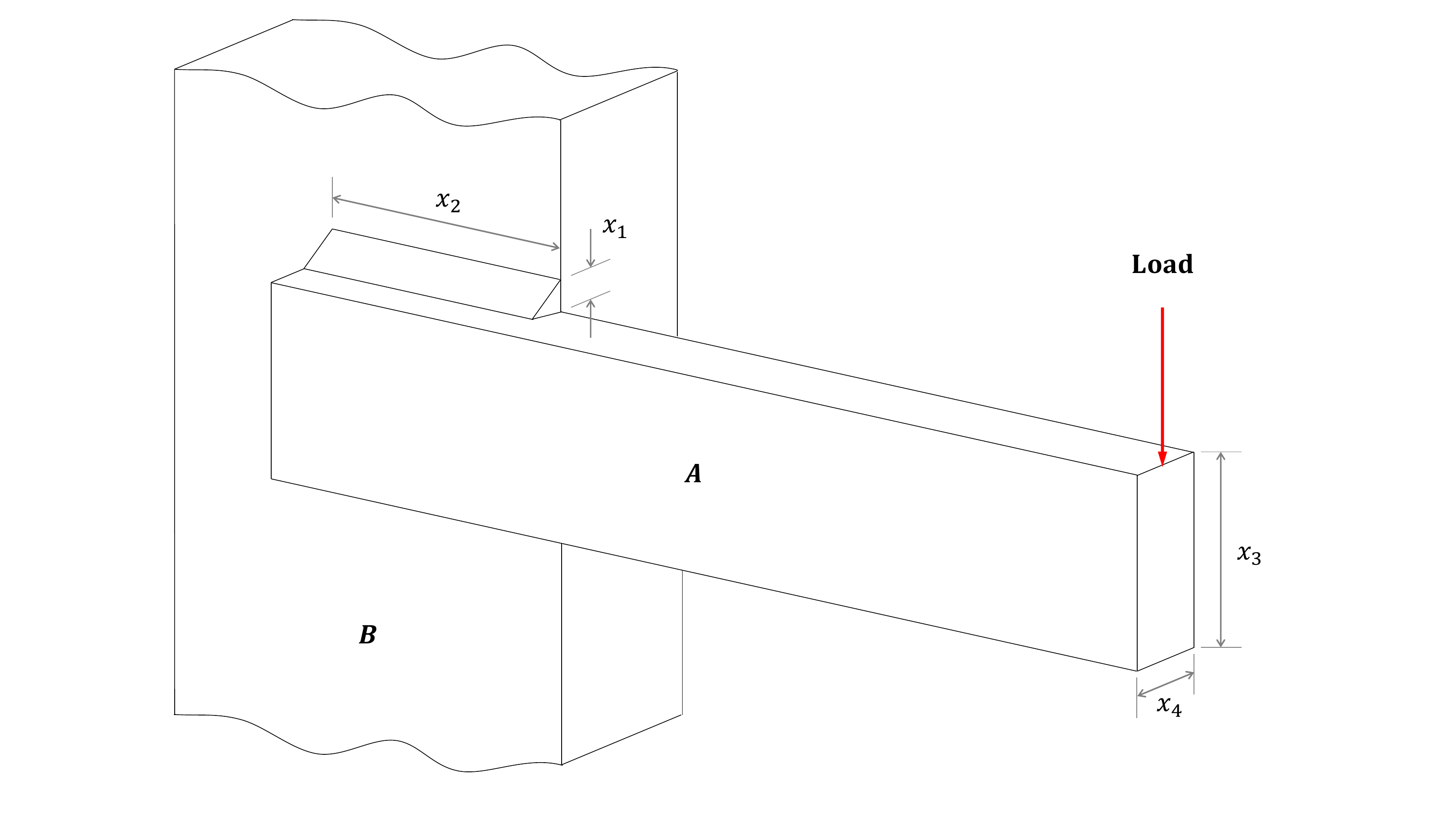}
\caption{Welded Beam.}
\label{fig:Welded_Beam}
\end{figure}
This second example concerns a welded beam structure, constituted of a beam A and the weld required to hold it to the member B, see \cref{fig:Welded_Beam}. The objective is to minimize its fabrication cost $f(X_1, X_2, X_3, X_4)$ under 5 nonlinear inequality constraints. The optimization is summarized in the equations below:
\begin{equation*}
f(X) = 1.10471X_1^2X_2 + 0.04811X_3X_4(14 + X_2) 
\end{equation*}
\noindent s.t. 
\begin{align*}
g_1(X) & = \tau(X) - 13600 \leq 0, \\
g_2(X) & = \sigma(X) - 30000 \leq 0, \\
g_3(X) & = X_1 - X_4 \leq 0, \\
g_4(X) & = 6000 - P_c(X) \leq 0, \\
g_5(X) & = \delta(X) - 0.25 \leq 0
\end{align*}
\begin{equation*}
0.125 \leq X_1 \leq 10, \; 
0.1 \leq X_2 \leq 10, \; 
0.1 \leq X_3 \leq 10, \;
0.1 \leq X_4 \leq 10
\end{equation*}
\noindent The terms $\tau(X)$, $\sigma(X)$, $P_c(X)$ and $\delta(X)$ are given in \Cref{app:WB4}.

\begin{figure}
\centering
\vspace{1cm}
\begin{tikzpicture}
\node[anchor=south west,inner sep=0] (image) at (0,0) {\includegraphics[width=\textwidth]{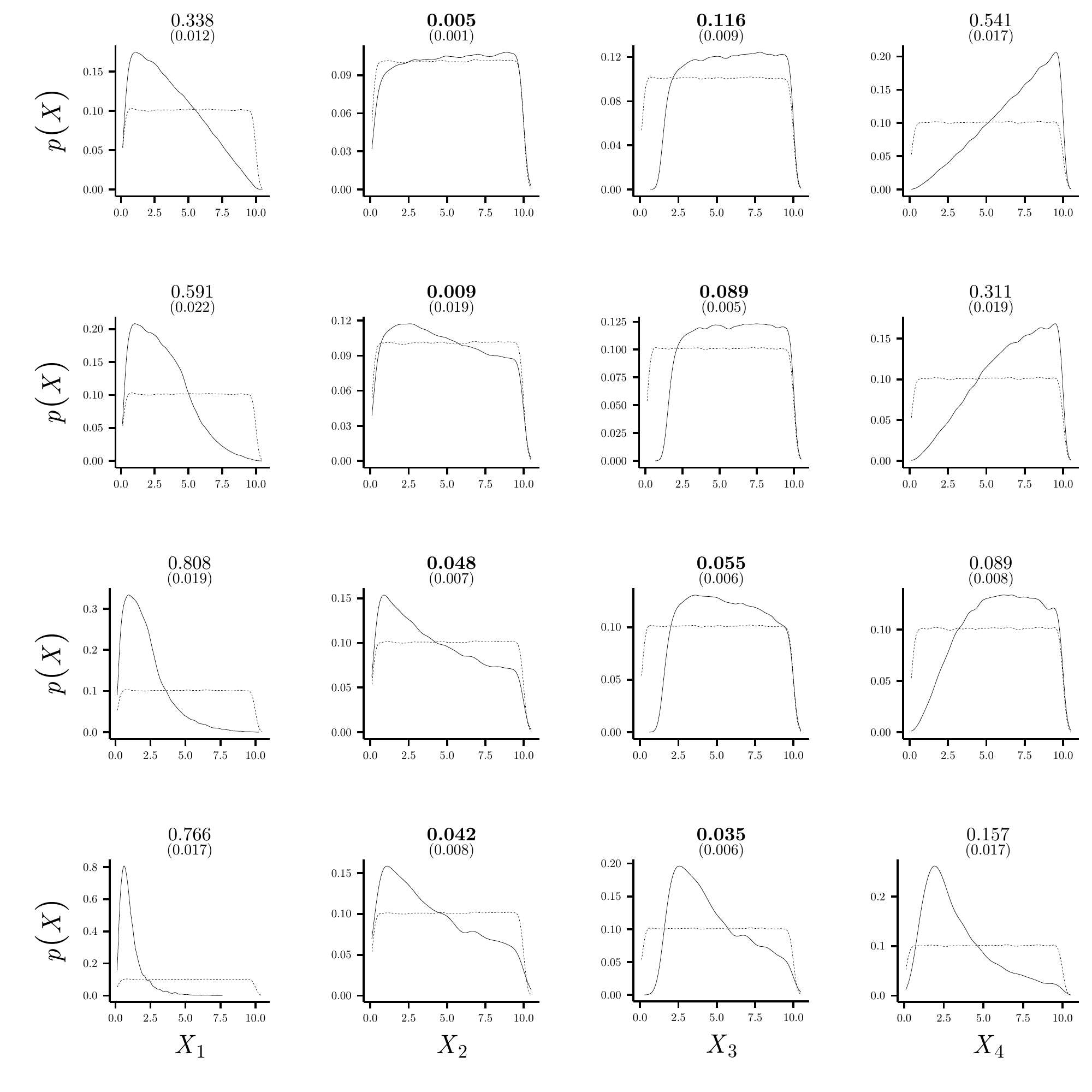}};
    \begin{scope}[x={(image.south east)},y={(image.north west)}]
        \node[inner sep=0pt] at (0.07,0.24){$\alpha = 10\%$};
		\node[inner sep=0pt] at (0.07,0.49){$\alpha = 40\%$};
		\node[inner sep=0pt] at (0.07,0.74){$\alpha = 70\%$};
		\node[inner sep=0pt] at (0.07,0.985){$\alpha = 100\%$};
    \end{scope}
\end{tikzpicture}
\caption{Evolution of $\mathbb{P}_{X_i | Z = 1}$ for different $\alpha$ value (continuous line) compared to the original distribution (dashed line) for the Welded Beam application. The continuous and dashed lines differ for $\alpha = 100\%$ because not all points are feasible. The values above each plot are the corresponding $\Shsici[q_\alpha,\textbf{T}]$ mean (and standard deviation). Bold numbers highlight small values of $\Shsici[q_\alpha,\textbf{T}]$ (and negligible $X_i$'s). For better readability, the scales of the vertical axes vary.} 
\label{fig:Dist_HSIC_WB4}
\end{figure}
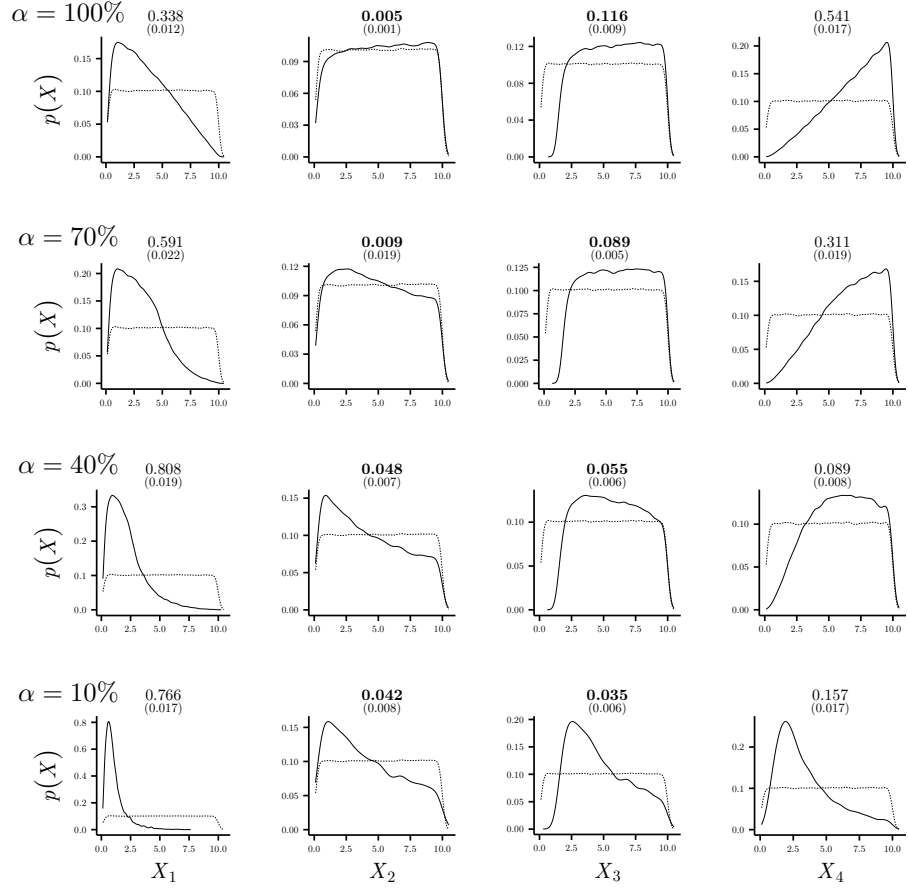

\paragraph{Sensitivity analysis} \cref{fig:Dist_HSIC_WB4} shows the evolution of the conditional distributions $\mathbb{P}_{X_i | Z = 1}$ for different quantiles $\alpha$. 
The mean and standard deviations of the HSIC-IT sensitivities associated to each $\alpha$, out of 20 repetitions, can be found above each plot. $X_2$ and $X_3$ are found to be negligible as their index is near zero for $\alpha = 10\%$. Their domains are only slightly restricted by the condition on performance, $Z=1$.

For the \emph{Greedy} problem modification, $X_2$ is set to 5.36 and $X_3$ to 8.54 as those values gave the lowest feasible objective function value during the sensitivity analysis. For reference, the optimal point found in the literature is $X^* = [0.206, 3.473, 9.037, 0.206]$.

\begin{figure} [p]
\centering
\scalebox{0.8}{
\begin{minipage}[c]{\linewidth}
\centering
\centerline{\includegraphics[scale = 0.6]{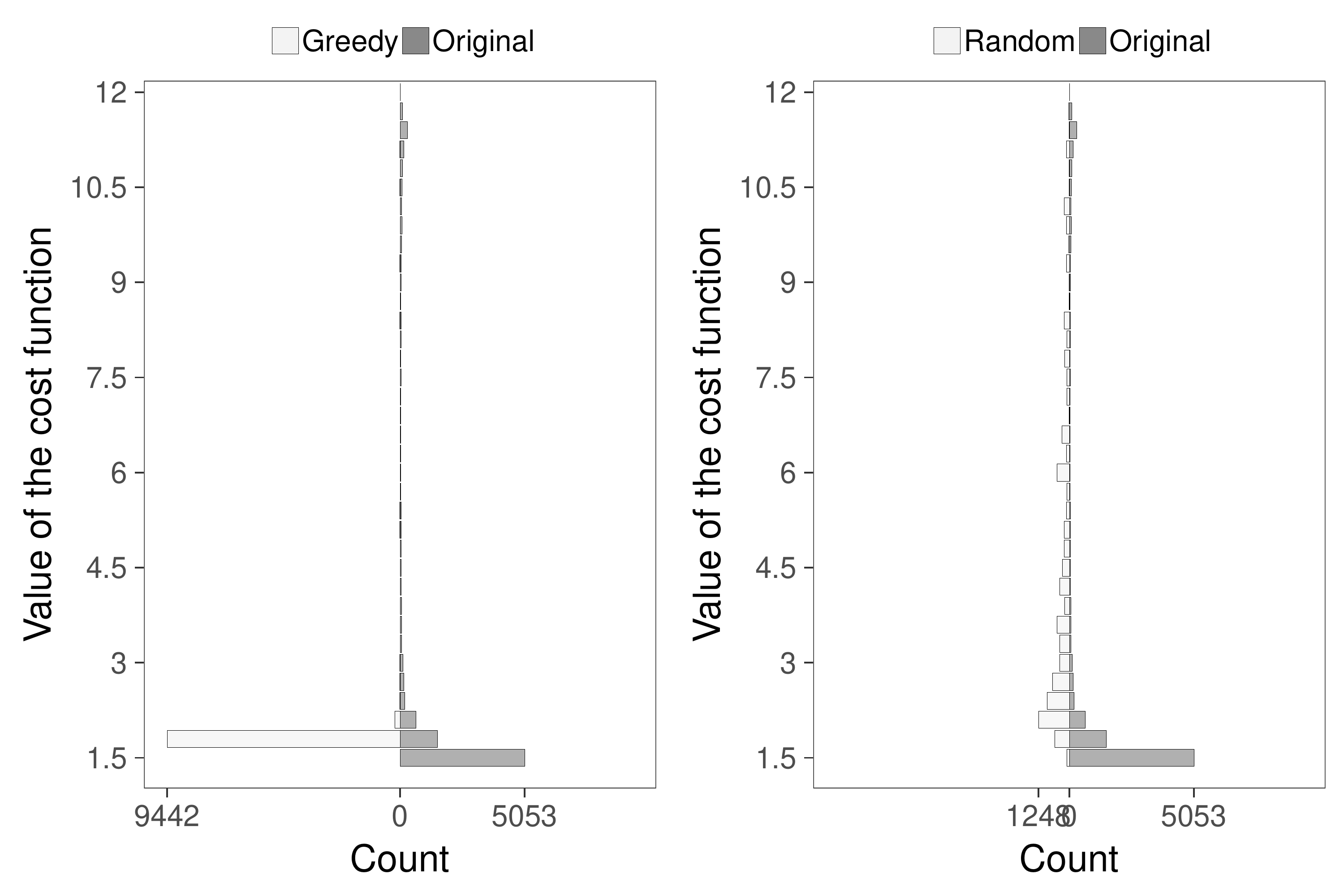}}
\centerline{\includegraphics[scale = 0.6]{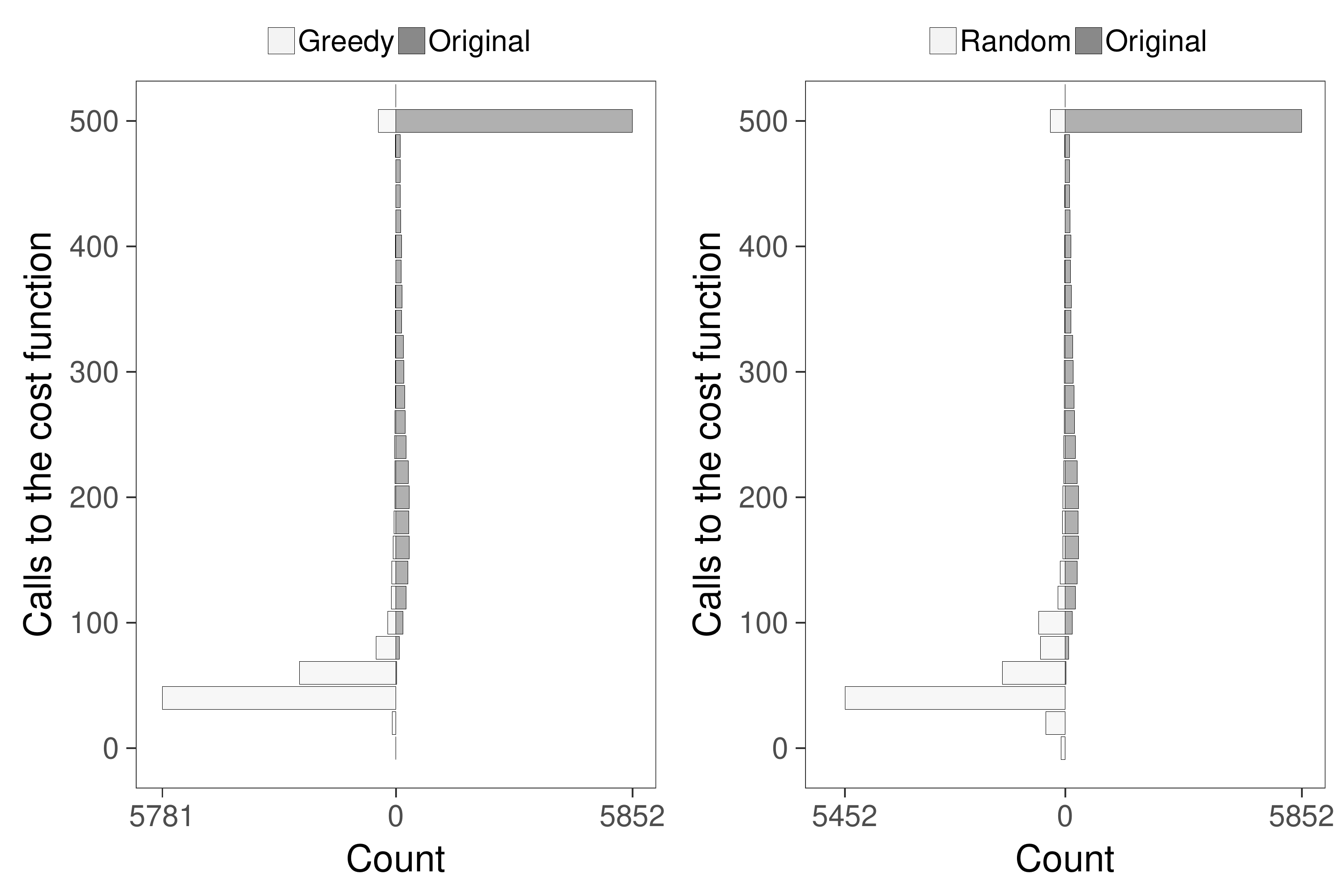}}
\end{minipage}}
\caption{Results of 10000 optimizations with the Original, Greedy and Random formulations  for the Welded Beam test case: histograms of the final objective functions (top) and number of calls to the objective function at convergence (bottom). }
\label{fig:results_optim_WB4}
\end{figure}

\paragraph{Optimization results} 
The 10000 optimizations of the WB4 problem are summarized in \cref{fig:results_optim_WB4}. It is seen on the top histograms that, with the original formulation, the global optimum of performance $f(X^*)=1.72$ is reached in half of the cases. However, as show by the bottom histograms, the associated number of calls to the objective function in almost all cases is the maximum budget (500). 
The \emph{Greedy} modification to the problem converges to a downgraded value of $f(X^*)=$1.97 at a much lower cost, with 50 calls to the objective function on the average. As can be observed on the top right histogram, the \emph{Random} modification to the problem yields inconsistent objective function values at convergence, because many choices of  ``negligible'' inputs lead to poor final achievable performance.

Once again, the freezing of some of the variables leads to savings in terms of calls to the objective function and to more robust convergences for the \emph{Greedy} version. Furthermore, it seems that the modified version no longer has the local optimum around $f(X^*)=11$ that is seen as a small mode in the top of \cref{fig:results_optim_WB4} in the original results.

\subsection{Discussion}
As seen in both above examples, setting variables with small \Shsici to a fixed value chosen with the \emph{Greedy} strategy led to significant improvements in terms of optimization cost and robustness, with an accompanying small degradation in performance at the optimum. This is due to the loss in fine-tuning ability resulting from freezing the value of the low impact inputs. This phenomenon was more visible with the \emph{Random} strategy where the variations in values of fixed inputs led to a spread in final objective functions.	
\\

\begin{figure} [h!]
  \begin{minipage}{0.5\linewidth}
	\centering
    \includegraphics[width=0.8\linewidth]{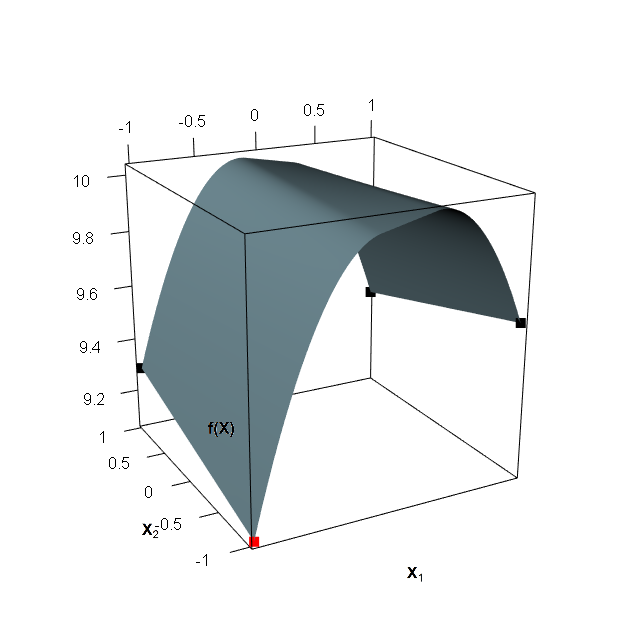}
  \end{minipage}%
  \begin{minipage}{0.5\linewidth}
	\centering
    \begin{minipage}{0.8\linewidth}
    \includegraphics[width=\linewidth]{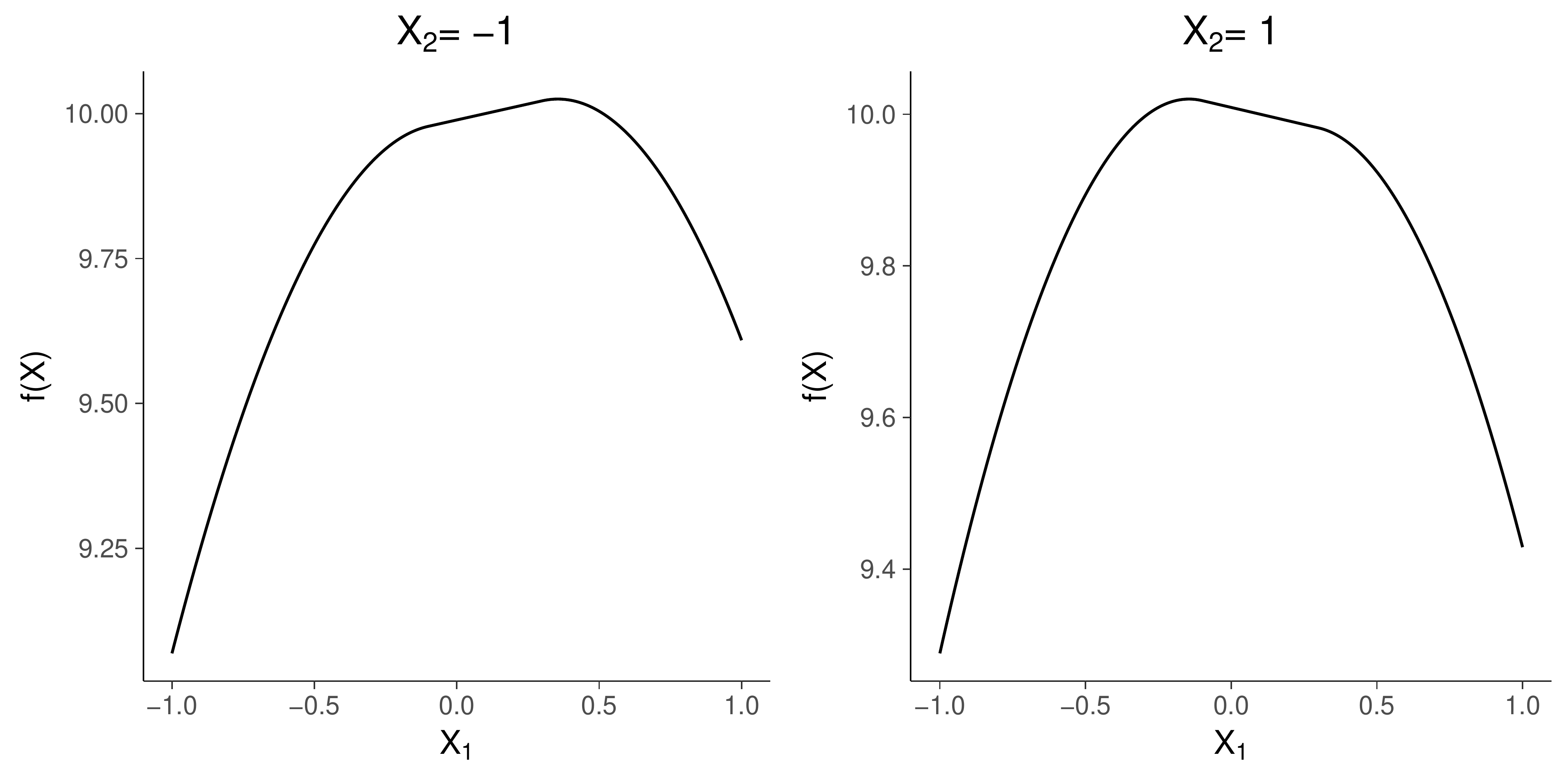}
    \end{minipage}
	\\[3mm]%
	\scalebox{0.9}{\begin{tabular}[t]{|c|c|c|}
	\hline 
	Original & $X_2 = -1$ & $X_2 = 1$ \\ 
	\hline 
	$55 \%$ & $67 \%$ & $42 \%$ \\ 
	\hline 
	\end{tabular}
	}
   \end{minipage}
\caption{\textbf{Left: } Surface plot of the twisted strip function with $\textbf{c} = (0.1,0.1)$, $A = 0.2$ and $\epsilon = 0.1$, \textbf{Upper-right: } Profile of the reduced objective function for different $X_2$ values, \textbf{Bottom-right: } Frequency of convergence to the global optimum for random initial $X_1$ depending of the value chosen for the frozen input $X_2$.}
\label{fig:DTS}
\end{figure}

We now argue, with the help of an illustrative example, that this impact is increased when the reduced problem is solved with a local optimization algorithm, such as in \Cref{sec:examples} (COBYLA was the local optimizer). Let us consider following two dimensional ``twisted strip'' toy function: 
\begin{equation*}
f(X) = 
 \begin{cases}
	10 - (\lvert X_1' \rvert - A)^2 - \epsilon X_2' X_1' &\quad \text{if} \quad \lvert X_1 \rvert \geq A \\
	10 - \epsilon X_2' X_1'  &\quad \text{otherwise}
 \end{cases}
\end{equation*}
with $X' = X - \textbf{c}$. The function is represented in \cref{fig:DTS} below for $\textbf{c} = (0.1,0.1)$, $A = 0.2$ and $\epsilon = 0.1$. This function possesses a global optimum at $(-1, -1)$ (the red square) and multiples local ones (the black squares), with a significant difference in the objective value. 

For this toy function, the HSIC-IT sensitivity index of the second variable is arbitrarily small, even for low quantiles. This can be seen by imagining the marginal distribution of $X_2$ when $f$ is restricted to low values, which is very close to the uniform distribution. Indeed, the twisted strip function is almost flat in the $X_2$ direction. Setting $X_2$ to a constant value simplifies the problem as it appears to be negligible. Whatever the chosen value of $X_2$, the reduced objective function has a global optimum at $X_1 = -1$ and a local one at $X_1 = 1$. The main difference between these 2 cases lies in the slope direction of the reduced function near $X_1 = 0 \pm A$, see the different profiles of the function in the upper-right side of \Cref{fig:DTS}. That implies that a local optimization algorithm will be sensitive to its initialization and will sometimes converge to the local optimum. Hence, depending of the choice made for the value of $X_2$, the frequency of convergence to the global optimum varies, increasing when $X_2$ is at its optimum (-1) and decreasing when away from it, see the table in \Cref{fig:DTS}. Such behaviour should be expected from functions with essential global optima, i.e., functions without ``needle in the haysack'', where the modified optimization problems lead to the global basin of attraction if the frozen variables are close to their optimum. In such well-behaved cases, the \emph{Greedy} heuristic gives $X_2 \approx -1$, leading to improved results. Note that the phenomenon of convergence to a local optimum and its dependency on the frozen variables would be much lessened if a global optimizer were used: the strategy proposed in this paper of HSIC-IT sensitivity analysis followed by a greedy freezing of some variables and an optimization would further benefit from a global optimizer.

Another bottleneck with highly constrained high dimension problems is finding feasible points. This requires to draw a large number of points to ensure that we have a large enough sample for the estimation of the sensitivity indices. To overcome this issue, we use a relaxation coefficient which correspond to $\textbf{T}$ in \cref{ssec:formulation} was made to get at least a hundred 100 feasible points for the computation of the HSIC-IT sensitivity index. 

Finally, in our example, we use a Gaussian RBF kernel for the inputs as a default choice but other kernels might be better suited for optimization variable selection. The heuristic choice of the bandwidth $\sigma$ might also be improved based on recent works that propose an optimization of the parameter, \cite{gretton2012optimal}.

\section{Conclusions}
\label{sec:conclusion}
This paper has studied how global sensitivity analysis can be specialized for contributing to the resolution of optimization problems. 

First, we have introduced three modifications of the objective function that are alternative expressions of the feasible level set idea. Each formulation is a different blend between two pieces of information, which inputs matter to reach an area close to the optima and how much each input impacts performance when being in such an area. The effect of each formulation on the Sobol indices has been observed. 

Second, building on the indicator-thresholding formulation in conjunction with the Hilbert Schmidt Independence Criterion, we have described a new HSIC-IT sensitivity index adapted to constrained optimization problems. This sensitivity index has been interpreted as a measure of the distance between two distributions, that of the variable being analyzed and that of the same variable conditional to its objective and constraints reaching a certain performance level.

Finally, the new HSIC-IT index has served to select variables before a local optimization is carried out. Provided that the variables which are not retained are given a value in a greedy manner, we have obtained in several test cases solutions with limited performance loss, at a substantially decreased number of function evaluations, and with more stable convergences.

In this work, the cost of calculating the HSIC-IT sensitivity indices has been left aside. An important, practical, perspective is to limit this cost by introducing statistical (surrogate) models in the HSIC-IT sensitivity estimation. 

\section*{Acknowledgments}
The authors would like to thank Olivier Roustant for his help with the project.

\bibliographystyle{siamplain}
\bibliography{references}
\appendix
\section{Welded Beam test case complementary expressions}
\label{app:WB4}

The expression of the terms $\tau(X)$, $\sigma(X)$, $P_c(X)$ and $\delta(X)$ is:

\begin{align*}
\tau(X)	& = \sqrt{\tau_1(X)^2 + \tau_2(X)^2 + X_2\tau_1(X)\tau_2(X)/ \sqrt{0.25(X_2^2 + (X_1 + X_3)^2)}}, \\
\sigma(X)	& = \frac{504000}{X_3^2X_4}, \\
P_c(X)		& = 102372.4(1 - 0.0282346X_3)X_3X_4^3, \\
\delta(X)	& = \frac{2.1952}{X_3^3X_4}, \\
\end{align*}
\noindent where 
\begin{align*}
\tau_1(X)	& = \frac{6000}{\sqrt{2}X_1X_2}, \\
\tau_2(X)	& = \frac{6000 (14 + 0.5 X_2) \sqrt{0.25\left( X_2^2 + (X_1 + X_3)^2\right)}}
					 {2\left( \sqrt{2}X_1X_2(X_2^2/12 + 0.25(X_1 + X_3)^2)\right)} ~.
\end{align*}

\end{document}